%% file: main.tex
\documentclass[sigconf]{acmart}
\AtBeginDocument{%
  }

\setcopyright{acmlicensed}
\copyrightyear{2025}
\acmYear{2025}
\acmDOI{XXXXXXX.XXXXXXX}
\acmConference[Conference acronym 'XX]{Make sure to enter the correct
  conference title from your rights confirmation email}{June 03--05,
  2018}{Woodstock, NY}
\acmISBN{978-1-4503-XXXX-X/2018/06}

\usepackage{enumitem}
 \usepackage{multirow}
\usepackage{subfigure}
\begin{document}

\title{From Noisy to Native: LLM-driven Graph Restoration for Test-Time Graph Domain Adaptation}


\author{Xiangwei Lv}
\authornote{Equal Contribution}
\email{xiangwei.lv@zju.edu.cn}
\affiliation{
    \institution{Zhejiang University}
    \city{Hangzhou}
    \country{China}}

\author{JinLuan Yang}
\authornotemark[1]
\email{yangjinluan@zju.edu.cn}
\affiliation{
    \institution{Zhejiang University}
    \city{Hangzhou}
    \country{China}}

\author{Wang Lin}
\email{linwanglw@zju.edu.cn}
\affiliation{
    \institution{Zhejiang University}
    \city{Hangzhou}
    \country{China}}

\author{Jingyuan Chen}
\authornote{Corresponding Author}
\email{jingyuanchen@zju.edu.cn}
\affiliation{
    \institution{Zhejiang University}
    \city{Hangzhou}
    \country{China}}

% \author{Mengze Li}
% \email{mengzeli@zju.edu.cn}
% \affiliation{
%     \institution{Zhejiang University}
%     \city{Hangzhou}
%     \country{China}}

% \author{Yongduo Sui}
% \email{yongduosui@tencent.com}
% \affiliation{
%     \institution{Tencent}
%     \city{Shenzhen}
%     \country{China}}

% \author{Zemin Liu}
% \email{liu.zemin@zju.edu.cn}
% \affiliation{
%     \institution{Zhejiang University}
%     \city{Hangzhou}
%     \country{China}}

\author{Beishui Liao}
\authornotemark[2]
\email{baiseliao@zju.edu.cn}
\affiliation{
    \institution{Zhejiang University}
    \city{Hangzhou}
    \country{China}}

\renewcommand{\shortauthors}{Xiangwei Lv et al.}
\def\method{GRAIL}
\begin{abstract}

% We view test-time adaptation as refine target graph to source characterisitics. Motivated by diffusion process, we ...

% Motivated by llm, llm 推理具备source style,we llm to reson graph token sequences and incorporate reinforce learning to guide the refine process with source property. In this way, we can utilize the finetuned llm to preprocess taregt graph to ensure it achiver superior performance when confronted domain shift. 

% We propose to utilize large language model to refine target graph data and enhance it with reinforce learning to ensure it with source characteristics. 
% Based on source data, we employ diffusion process to construct noisy state and use them to fine-tune a large language model. In this way, the language can understand graph sequence and performance graph refinement inference. Further, the reinforce learning is proposed to improve refinement quality through source characteristics, like degree distribution. Extensive experiment demonstrate our method's effectiveness. 

Graph domain adaptation (GDA) has achieved great attention due to its effectiveness in addressing the domain shift between train and test data. A significant bottleneck in existing graph domain adaptation methods is their reliance on source-domain data, which is often unavailable due to privacy or security concerns. This limitation has driven the development of Test-Time Graph Domain Adaptation (TT-GDA), which aims to transfer knowledge without accessing the source examples. Inspired by the generative power of large language models (LLMs), we introduce a novel framework that reframes TT-GDA as a generative graph restoration problem, \textit{"restoring the target graph to its pristine, source-domain-like state"}. There are two key challenges: (1) We need to construct a reasonable graph restoration process and design an effective encoding scheme that an LLM can understand, bridging the modality gap. (2) We need to devise a mechanism to ensure the restored graph acquires the intrinsic features of the source domain, even without access to the source data. 
To ensure the effectiveness of graph restoration, we propose a new approach, named \method{}, that restores the target graph into a state that is well-aligned with the source domain. Specifically, we first compress the node representations into compact latent features and then use a graph diffusion process to model the graph restoration process. Then a quantization module encodes the restored features into discrete tokens. Building on this, an LLM is fine-tuned as a generative restorer to transform a ``noisy'' target graph into a ``native'' one. To further improve restoration quality, we introduce a reinforcement learning (RL) process guided by specialized alignment and confidence rewards. This ensures the restored target graph possesses the key attributes of the source graph.
Extensive experiments demonstrate the effectiveness of our approach across various datasets. 

% Our code is vailable at \textcolor{blue}{\url{https://anonymous.4open.science/r/GRAIL-507D}}.

% reward signal: 
% feature, structure 

\end{abstract}




\received{20 February 2007}
\received[revised]{12 March 2009}
\received[accepted]{5 June 2009}

\maketitle

\input{introduction}

\section{Related Work}

Graph Domain Adaptation (\textbf{GDA}) aims to effectively transfer knowledge from a labeled source graph to an unlabeled target graph \cite{shen2023domain, zhang2024collaborate, wu2024graph, liu2024beyond}. Most existing GDA approaches address this challenge through adversarial learning \cite{udagcn, shen2023domain, mancini2019adagraph} or by minimizing the distributional distance \cite{mmd, yan2017mind} between the source and target graphs, often using metrics like Maximum Mean Discrepancy (MMD). However, these methods heavily rely on direct access to the source graph, which is often infeasible in real-world applications due to privacy, security, or data ownership constraints.
% Graph Domain Adaptation \textbf{GDA} aims to effectively transfer knowledge learned from a labeled source graph an unlabeled target graph. Most existing approchs attempt to address it through adversarial learning \cite{udagcn} or minimizing the distance \cite{mmd} between source and target graphs. However, these appraochs heavily rely on direct access to the source graph, which is usually not accessible in real-world scenarios due to privacy or safety concerns. 

To overcome the limitations of traditional GDA, Test-Time Graph Domain Adaptation (\textbf{TT-GDA}) has emerged as a promising research area. While test-time adaptation (TTA) methods have been explored in the vision domain \cite{ma2024improved, tomar2023tesla, liang2025comprehensive}, applying them to graph data presents unique challenges due to its complex topological structures and relational dependencies. 
Existing TT-GDA approaches can be broadly categorized into two main groups based on how they adapt the pre-trained model on source graphs: (1) Data-Centric Approaches: These methods primarily focus on modifying or augmenting the input data to facilitate adaptation. For instance, GTrans \cite{gtrans} uses a surrogate loss to guide the adaptation process, while GraphPatcher \cite{graphpatcher} improves predictions on low-degree nodes by introducing virtual nodes connected to the existing graph. (2) Model-Centric Approaches: These approaches directly adjust the learned model's parameters to align with the target domain. For example, SOGA \cite{soga} utilizes mutual information maximization and a structure consistency objective to improve the adaptation process. Besides, Matcha \cite{matcha} proposes adjusting the hop-aggregation parameters of the graph neural network to mitigate the effects of structural domain shifts. Our approach follows the data-centric perspective and creatively uses an LLM to restore the target graph with source characteristics, allowing it to perform well with the pre-trained source model.

% To address it, Test-Time Graph Domain Adaptaion \textbf{TT-GDA} is proposed and has attracted researchers's interest. Although there exits several TTA works on vsison domain, graph data's complex topologies make simply transfer unsatisfactory, which makd TT-GDA is a promising and challenging task. We split the existing TT-GDA approach into two parts based on whether change the learned model paramters on source graph: (1) data-centric approach and (2) model-centraic appraoch. For data-centric appraoch, GTrans \cite{gtrans} employ a surrogate loss to guide the graph adaptation. GraphPatcher utilize  virual nodes connetected to ecisting graph nodes to ompreovr the prediction on low-degree nodes. 
% For model-centric appraoch, SOGA \cite{soga} introduce mutual inforamtion maximization and structure consistency to improve adaptation process. Matcha \cite{matcha} prose adjusting the hop-aggregation parameters to amitigating structure shifts. 
\section{Preliminaries}
% Given a graph $G = (V, E)$ where $V$ is node set with $n = \vert V \vert$ nodes and $E$ denotes the edge set with $m = |E|$ edges. The adjency matrix $A \in \mathbb{R}^{n \times n}$ with $A_{ij} = 1$ means there is an edge between node $v_i \in V$ and $v_j \in V$, $A_{ij}=0$ otherwise. 
% We define node feature matrix as $X \in \mathbb{R}^{n \times d}$. Here $d$ denotes the dimension of feature matrix. Here we use $G^s$ and $G^t$ to represent source graph and target graph respectively. The source graph with label matrix $Y \in \{0, 1\}^{n \times C}$, where $C$ is the number of labels. Let $f_{GNN}$ represent a Graph Neural Network trained on source graph on node classfication task. The goal of test-time graph domain adaptation (TT-GDA) aims to perform well on target graph $G^t$ through utilizing $f_{GNN}$ pretrained on source graph, meanwhile source graph $G^s$ is not allowed during adaption during adaptation. 

% In this work, we focus on adpatation on node classification. The source graph $G^s$ is accompanied by a label matrix $Y \in \{0, 1\}^{n \times C}$, where $C$ is the number of classes.

\subsection{Problem Definition} Given a graph $G = (V, E)$, where $V$ is the set of nodes and $E$ is the set of edges. Let $n = |V|$  and $m = |E|$ denote the total number of nodes and edges, respectively. The graph structure is represented by an adjacency matrix $A \in \{0, 1\}^{n \times n}$, where $A_{ij} = 1$ if an edge exists between node $v_i \in V$ and $v_j \in V$, and $A_{ij}=0$ otherwise. The feature matrix is represented as $X \in \mathbb{R}^{n \times d}$, where $d$ is the feature dimension. In the context of domain adaptation, we distinguish between a source graph $G^s$ and a target graph $G^t$. They share the same label space but have different data distributions.
The source graph $G^s$ is accompanied by a label matrix $Y \in \{0, 1\}^{n \times C}$, where $C$ is the number of classes.
We assume a Graph Neural Network (GNN), denoted as $f^s_{GNN}$, has been pre-trained on the source graph $G^s$ for a node classification task.
The objective of \textbf{Test-Time Graph Domain Adaptation (TT-GDA)} is to effectively improve the performance of the pre-trained model $f^s_{GNN}$ on the target graph $G^t$. A critical constraint is that the source graph $G^s$ is not accessible during the adaptation process, typically due to privacy concerns.

\subsection{Graph Neural Networks} 
Graph Neural Networks (\textbf{GNNs}) are models designed to process graph data by iteratively aggregating and transforming information from a node's neighbors \cite{gcn, gat, gin, gat}. This message-passing mechanism updates the feature representation of each node, enabling GNNs to learn powerful node embeddings. A typical GNN layer can be expressed as:
\begin{equation}
    h_i^{(l+1)} = \text{UPDATE}^{(l)} \left( h_i^{(l)}, \text{AGG}^{(l)} \left( \{h_j^{(l)} | v_j \in \mathcal{N}(v_i)\} \right) \right)
\end{equation}

where $h_i^{(l)}$ is the feature vector of node $v_i$ at layer $l$, and $\mathcal{N}(v_i)$ is the set of neighbors of node $v_i$. $\text{AGG}^{(l)}$ denote the aggragation function to aggregates the feature vectors from the neighbors. And $\text{UPDATE}^{(l)}$ represnt the update to combines the aggregated neighbor information with the node's current features to generate a new representation. After $L$ layers, a final set of node embeddings is obtained for tasks like node classification. 

\subsection{Diffusion Models}
Diffusion models have become a prominent class of generative models, achieving significant progress in computer vision and other domains \cite{yang2023diffusion, nichol2021improved,ozbey2023unsupervised}. They are primarily composed of two core phases: a forward diffusion process and a reverse denoising process. 
% The former progressively adds noise to data, while the latter learns to reverse this process to generate new samples.

Given a data distribution $p(\mathbf{x}_0)$, the forward process gradually adds Gaussian noise to the data. This is modeled as a Markov chain where, at each step $t$, the transition from  $x_{t-1}$ to $x_t$ is defined by the following distribution:
\begin{equation}
q(\mathbf{x}_t|\mathbf{x}_{t-1}) = \mathcal{N}(\mathbf{x}_t; \sqrt{1 - \beta_t }\mathbf{x}_{t-1}, \beta_t\mathbf{I})
\end{equation}
Here, $\beta_t \in (0, 1)$ is a predefined variance schedule that controls the amount of noise added at step $t$, and $\mathbf{I}$ is the identity matrix. A key property of this process is that we can directly sample $\mathbf{x}_t$ from the initial data point $\mathbf{x}_0$ without needing the intermediate steps. This is made possible by the reparameterization trick, which yields the following relationship:
\begin{equation}
\mathbf{x}_t = \sqrt{\bar{\alpha}_t}\mathbf{x}_0 + \sqrt{1 - \bar{\alpha}_t}\mathbf{\epsilon_t}
\end{equation}
where $\alpha_t = 1 - \beta_t$, $\bar{\alpha}_t = \prod_{i=1}^{t}\alpha_i$, and $\mathbf{\epsilon_t} \sim \mathcal{N}(\mathbf{0}, \mathbf{I})$ is a standard Gaussian noise vector.

The reverse process aims to train a neural network, parameterized by $\theta$, to model the reverse transitions $p_{\theta}(\mathbf{x}_{t-1}|\mathbf{x}_t)$. This learned process progressively removes the noise added in the forward process to generate new data. Similar to the forward transitions, the reverse transition is also modeled as a Gaussian distribution:
\begin{equation}
p_{\theta}(\mathbf{x}_{t-1}|\mathbf{x}_t) = \mathcal{N}(\mathbf{x}_{t-1};\mathbf{\mu}_{\theta}(\mathbf{x}_t, t), \mathbf{\Sigma}_{\theta}(\mathbf{x}_t, t))
\end{equation}
where $\mathbf{\mu}_{\theta}(\mathbf{x}_t, t)$ and $\mathbf{\Sigma}_{\theta}(\mathbf{x}_t, t)$ represent the predicted mean and covariance of the reverse distribution, respectively. In the Denoising Diffusion Probabilistic Models (DDPM) framework \cite{ho2020denoising},  the covariance is often fixed to a small constant, such as $\mathbf{\Sigma}_{\theta}(\mathbf{x}_t, t)=\beta_t\mathbf{I}$. The $\mathbf{\mu}_{\theta}(\mathbf{x}_t, t)$ is learned by the neural network, can be derived through reparameterization trick as:
\begin{equation}
\mathbf{\mu}_{\theta}(\mathbf{x}_t, t) = \frac{1}{\sqrt{\alpha_t}}\left(\mathbf{x}_t - \frac{\beta_t}{\sqrt{1-\bar{\alpha}_t}}\mathbf{\epsilon}_{\theta}(\mathbf{x}_t, t)\right)
\end{equation}
where $\mathbf{\epsilon}_{\theta}(\mathbf{x}_t, t)$ denote the noise prediction network, which can be implemented through neural networks. Finally, the mean squared error is used as objective loss to train $\mathbf{\epsilon}_{\theta}(\mathbf{x}_t, t)$ by:
\begin{equation}
\label{loss:diffusion}
    \mathcal{L} = \mathbb{E}_{t , \mathbf{x}_0, \mathbf{\epsilon_t}} \left[ \left\| \mathbf{\epsilon}_t - \mathbf{\epsilon}_{\theta}(\mathbf{x}_t, t) \right\|^2 \right]
\end{equation}

\input{method}
\input{experiments}

% codebook number
% diffusion steps

\begin{figure}[t]
    \centering
    \subfigure[$\lambda_1$]{
    \includegraphics[width=0.47\linewidth]{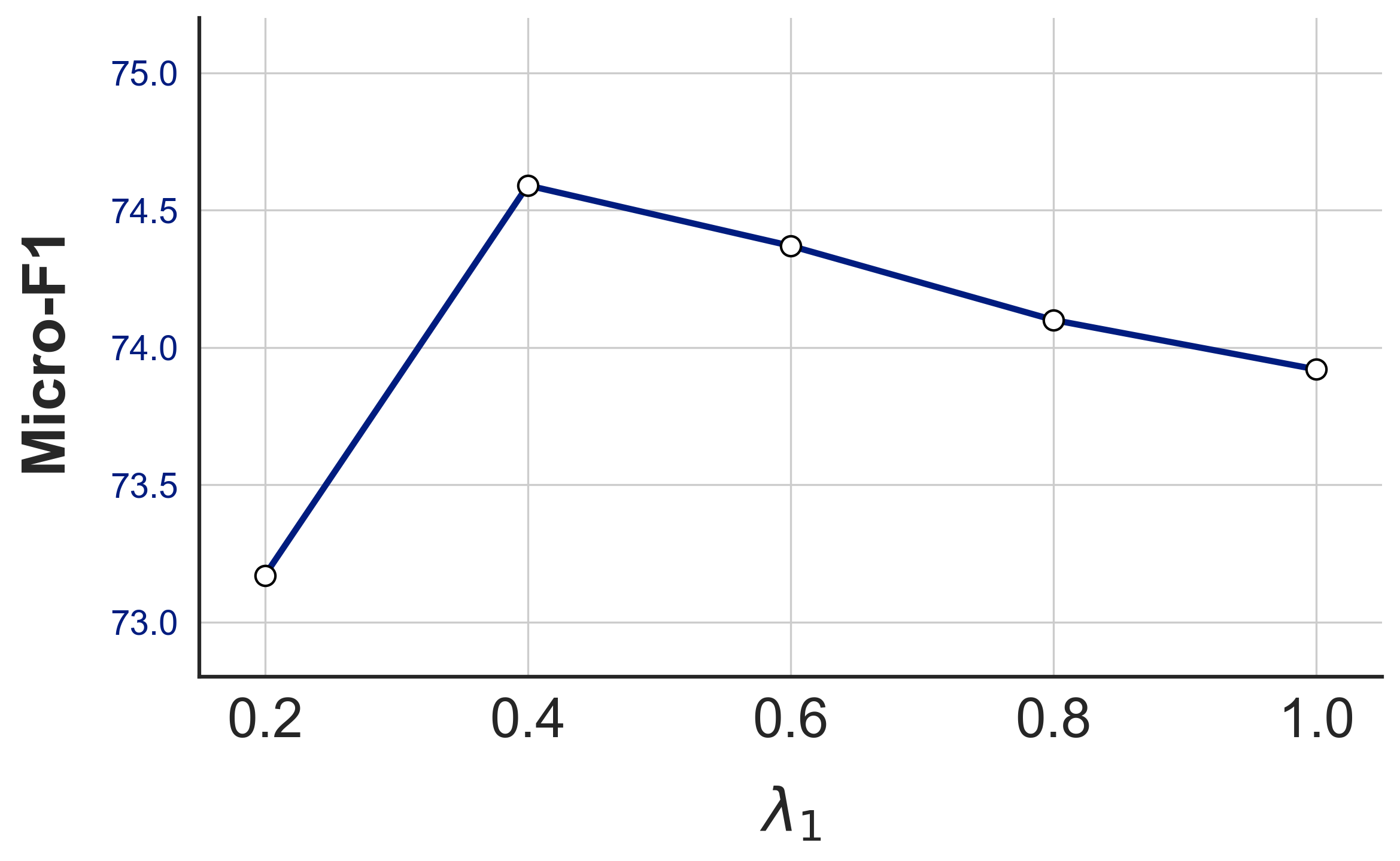}
    \label{fig:vis:before_class}
    }
    \subfigure[$\lambda_2$]{
    \includegraphics[width=0.47\linewidth]{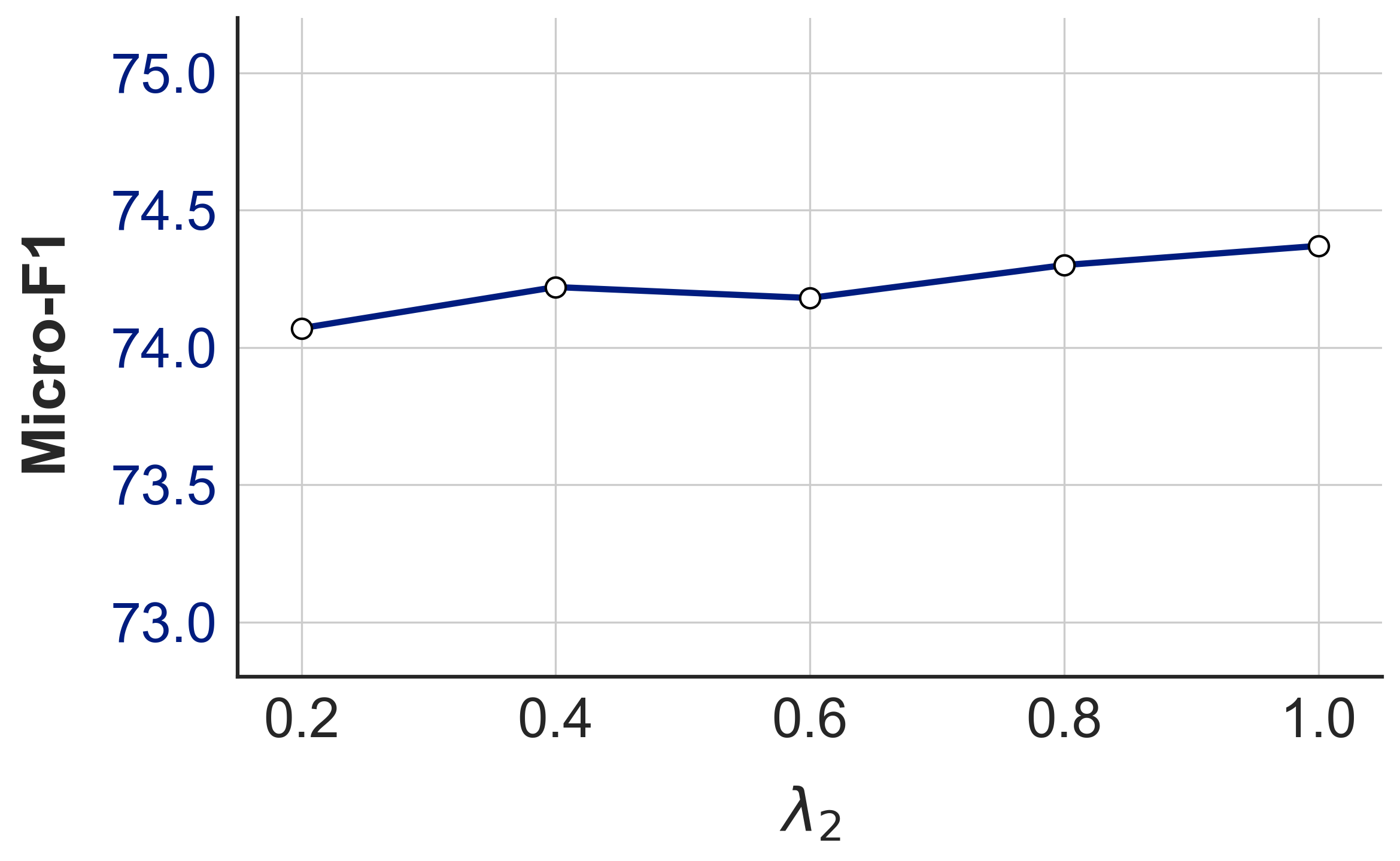}
    \label{fig:vis:after_class}
    }
    \vspace{-1em}
    \caption{Micro-F1 of various weight parameters.}
    \vspace{-1em}
    \label{fig:weight}
\end{figure}

\section{Conclusion}
This paper introduces GRAIL, a novel framework for Test-time Graph Domain Adaptation (TT-GDA). Addressing a key limitation of traditional methods, GRAIL leverages a fine-tuned Large Language Model (LLM) to refine target graphs without access to source data. Our core contributions lie in solving two fundamental challenges: bridging the modality gap between graphs and text, and providing effective source-free guidance.
To bridge the modality gap, we propose a novel graph diffusion-based tokenization scheme, which transforms complex graph structures into learnable discrete tokens for the LLM. For source-free guidance, we introduce a reinforcement learning algorithm with specialized alignment and confidence rewards, ensuring the refined graphs align with the target domain's latent characteristics.
Comprehensive experiments on various datasets validate GRAIL's effectiveness, demonstrating its superior performance over state-of-the-art methods.

% \clearpage
\bibliographystyle{ACM-Reference-Format}
\bibliography{main}

% \clearpage
% \appendix


\end{document}

%% file: introduction.tex
\section{Introduction}

% challenge
% 1. Tradition data-centric method 
% 2. 

% method 
% 1. 
% 2. 

% Recent the development of large lanaguage models have demonstrated their strong resoning capabilities. It motivate us to effectively utilize to guide the graph restoration process. 
% To address it, two key issues exitss: (1) how to transform the graph restoration process to a form LLM can understand. Here includes how to enocder the graph data and how to build efficient restoration process. (2) Without source data, how to impose engough inforamtion on taregt graph refine prrocess to ensure them with source characteristics, especially under LLM settsing.
% To address them, we propose GRAIL approach address the challenge of using LLM to conduct target graph restoration process with source charactersic.  

Graph Neural Networks (GNNs) have demonstrated remarkable success in modeling structured data across diverse applications, including social networks \cite{davies2022realistic, fan2019graph}, citation networks \cite{gcn}, and so on. However, their efficacy relies on the critical assumption that training and test data are independently and identically distributed (i.i.d.), which is rarely met in real-world applications. When a GNN is trained on a specific source domain and deployed to a different target domain, it often experiences a significant performance drop due to domain shift. This shift can be attributed to changes in both feature distributions and graph topology \cite{udagcn}. For example, a GNN trained on a citation graph of older ArXiv papers may struggle with more recent ones \cite{soga}. This is because shifts in research interests alter the feature distribution, while the emergence of new fields like deep learning profoundly evolves the graph's topology.

To mitigate the above problem, \textbf{Graph Domain Adaptation (GDA)} has emerged, aiming to transfer knowledge from a labeled source domain to an unlabeled target domain \cite{wu2024graph}. Most existing GDA methods \cite{acdne, udagcn, mfrreg, adagcn, semigcl} align the feature spaces of the two domains by relying on direct access to source data. However, this is often impractical due to privacy or intellectual property constraints \cite{soga, graphpatcher, gtrans, matcha}. This practical barrier has motivated a promising research direction: \textbf{Test-time Graph Domain Adaptation (TT-GDA)}, which aims to adapt a pre-trained source model to a target graph at test time, without access to the original source data.  
Inspired by the success of Large Language Models (LLMs) in text, vision, and multimodal domains \cite{gpt4, blip2, sora}, we naturally consider how to leverage their powerful generative and reasoning capabilities \cite{ds-r1, li2025system} to facilitate the graph adaptation process. Our core idea is to utilize an LLM to learn the latent structural and stylistic properties of a source domain. This implicit knowledge is then applied to refine a target graph, effectively bridging the domain gap without needing direct access to the source data.
However, developing an LLM-driven TT-GDA framework presents two fundamental challenges: (1) \textbf{Graph Restoration-based Tokenization}. The first challenge is the fundamental modality gap between the non-Euclidean, topological nature of graphs and the sequential, text-based architecture of LLMs \cite{fatemi2023talk, ren2024survey}. Simply serializing a graph's nodes and edges into a list is a naive solution that fails to capture the rich, high-order structural information and permutation invariance inherent to graph data. Therefore, the core challenge is to devise a effective tokenization scheme that can transform a graph into discrete semantic tokens, where the sequence itself reflects a valid graph formation or restoration process. (2) \textbf{Effective Alignment with Source Characteristics.} The second challenge is guiding the LLM's generation process to faithfully instill the target graph with the latent characteristics of an unseen source domain. It is impractical to deterministically predefine an optimal, source-styled restoration process for target graphs. A straightforward solution is to use the source graph to create pseudo graph restoration sequences for supervised fine-tuning (SFT) of an LLM. However, the LLM trained solely via SFT may not effectively determine if its modifications truly conform to the source style, making it a significant challenge to guarantee the quality of the refined target graph.  

% To address these challenges, we propose a novel framework, called \textbf{\method{}}, which leverages a fine-tuned LLM to perform TT-GDA task through target graph restoration and alignment. 
To address these challenges, we propose \textbf{\method{}}, a novel framework that leverages a fine-tuned LLM to perform the TT-GDA task via target graph restoration and alignment.
The overall approach consists of two phases: (1) Graph Diffusion Trajectory Tokenizer and (2) LLM-based Graph Restoration and Alignment. In phase 1, we first utilize a Q-former-based encoder compresses the variable-sized input graph into a set of fixed and compact latent representations. Then, a graph diffusion model learns the generative process of graph restoration at this latent level, producing a sequence that mirrors the step-by-step restoration of a graph. These continuous latent vectors are then discretized into token IDs via a vector quantizer, preparing the data for the LLM.
In phase 2, we fine-tune an LLM on token sequences generated by phase 1 to master the task of graph restoration. To enhance the quality of target graph refinement without source data, we introduce a reinforcement learning algorithm with specialized alignment and confidence rewards, aiming to encourage the LLM to produce a graph that is well-aligned with the latent characteristics of the source domain.

Our contributions are summarized as follows:
\begin{itemize}[leftmargin=*]
    \item We are the first to formalize and address the problem of Test-Time Graph Domain Adaptation (TT-GDA) by using a Large Language Model to refine the target graph data, eliminating the need for source data during adaptation.
    \item We propose GRAIL, a novel framework comprising two key components: a graph diffusion trajectory tokenizer that models the graph restoration process and then tokenizes its trajectory for bridging the modality gap, and a reinforcement learning algorithm that provides effective source-free guidance for graph refinement.
    \item We conducted comprehensive experiments on multiple benchmarks, which validate the effectiveness of GRAIL by showing it significantly outperforms existing state-of-the-art methods.
\end{itemize}

%% file: method.tex
\section{Method}
\subsection{Overview}

\begin{figure*}
    \centering
 \includegraphics[width=\linewidth]{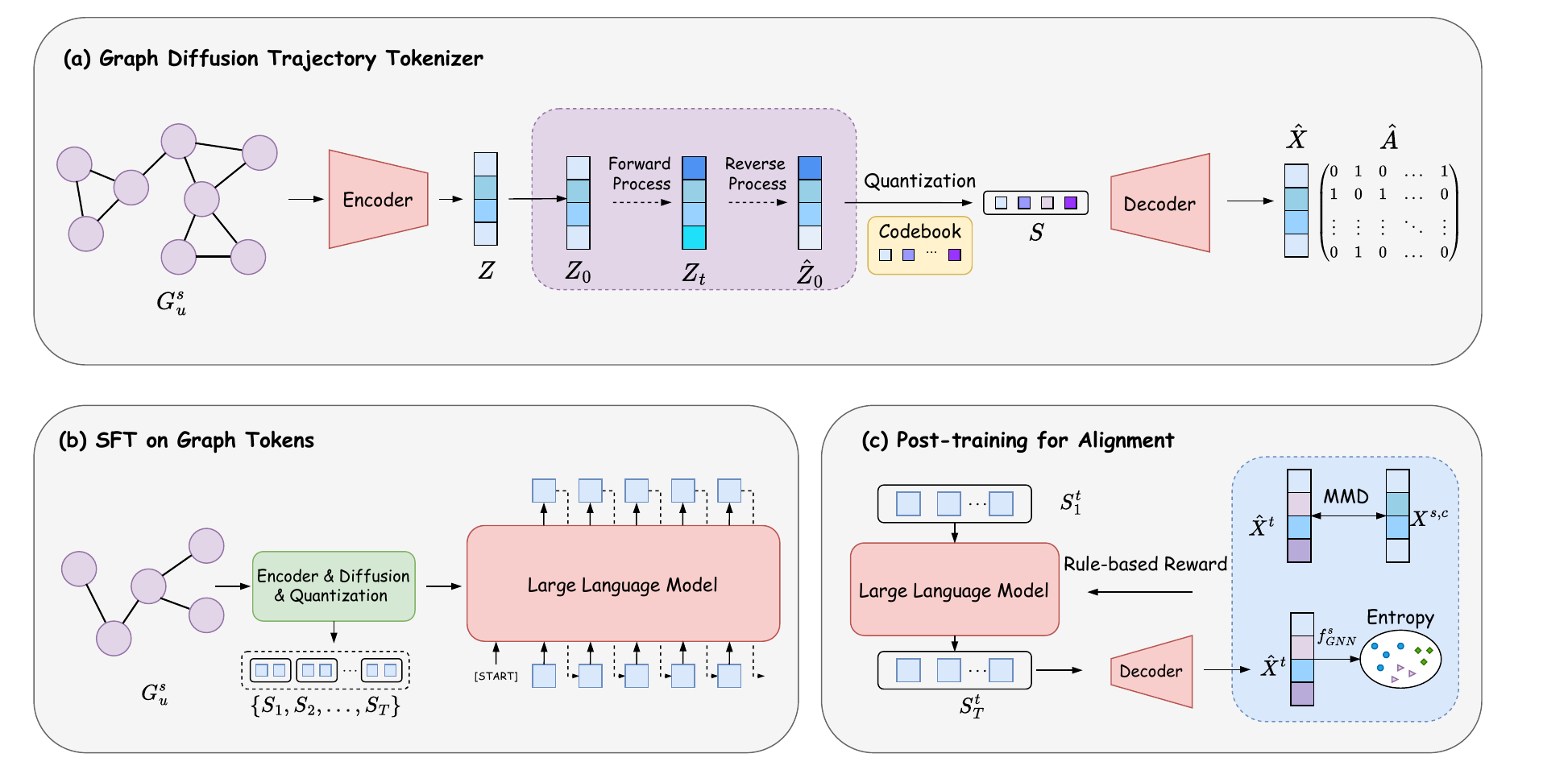}
    % \vspace{-0.3cm}
    \caption{The architecture of \method{}. (a) \textit{Graph Diffusion Trajectory Tokenizer} utilizes a graph diffusion process to create a source graph restoration trajectory at the embedding level. The trajectory is subsequently tokenized into discrete token IDs. An additional decoder then reconstructs the graph from these tokenized ids. (b) \textit{SFT on Graph Tokens} uses the tokenized sequences from stage (a) to fine-tune an LLM, which enables the LLM to understand and model the graph restoration process in an autoregressive manner. (c)  \textit{Post-training for Alignmnet} further enhances the LLM's restoration capabilities by incorporating a reinforcement learning process, which leverages alignment and confidence rewards to refine target graphs with source characteristics.}
    \label{fig:framework}
    % \vspace{-0.3cm}
\end{figure*}

In this section, we propose \method{}, a novel method that leverages Large Language Models (LLMs) to perform graph restoration and effectively align a target graph with source-domain characteristics. As illustrated in Figure~\ref{fig:framework}, our approach comprises two key stages: (1) \textbf{Graph Diffusion Trajectory Tokenizer}, (2) \textbf{LLM-based Graph Restoration and Alignment}.
% In this section, we propose a novel method, called \method{}, which aims utilize LLM to perform graph restoration process and effectively align target graph with source characteristics. To address it, as shown in Figure\ref{fig:framework}, our approach consists of following three stages: (1) \textbf{Graph Diffsuion Tokens Construction}. (2) \textbf{Pre-Training for Graph Tokens}. (3) \textbf{Post-Trainging for Source Characteristics Alignment}. 
% \subsection{Encoder}. 
In \textbf{stage 1}, we aim to obtain desired restored graph sequences and effectively encode them into token IDs for subsequent LLM training. To achieve this, we first introduce a Q-Former framework to compress a variable number of nodes into a fixed and compact set of latent representations. A diffusion model is then trained on these latent features through a forward noise-adding process and a reverse denoising process. To further map these latent representations into tokens, we incorporate a quantizer to discretize the continuous representations. Finally, a decoder is employed to reconstruct the graph data to mitigate information loss. 
In \textbf{stage 2}, We first leverage the diffusion model trained in Stage 1 to generate graph sequences that simulate the graph restoration process. These generated sequences are subsequently tokenized into discrete token streams. We then employ an LLM to model this restoration process in an autoregressive manner. To overcome the limitations of simple imitation learning, we further enhance the model through a reinforcement learning process. This stage guides the LLM to discover implicit source characteristics and perform graph refinements by using a dual-component reward function, which includes an alignment reward and a confidence reward.

\subsection{Graph Diffusion Trajectory Tokenizer}

% To effectively model the graph restoration process, we employ diffusion models to generate the denoising procedure. 

As mentioned, our aim is to model the evolution of graph restoration to capture the inherent source characteristics so that we can refine the target graph for improved adaptation. Due to the complex and variable topology of graphs, we propose to first encodes them into manageable embeddings, which then undergo a diffusion process. These embeddings can then transformed into discrete tokens to serve as the basic units for training an LLM. Since our primary focus is on node classification, we use the subgraph $G^s_u$ centered at node $u \in V$ as our initial input. We detail more components below.
% As mentioned, our goal is to model the evolution of graph restoration. By capturing the inherent source characteristics, we can refine the target graph to achieve better adaptation performance. However, due to the complex topology of graphs, such as a variable number of nodes, we propose an approach that first encodes the graphs into manageable embeddings. These embeddings then undergo a diffusion process. Furthermore, the embeddings can be transformed into discrete tokens, which serve as the basic units for subsequent large language model (LLM) training.
% Since our focus is mainly on node classification, we use subgraph $G^s_u$ centered at node $u$ as initial input. 

% However, for node classification tasks, the number of nodes and the adjacency matrix are variable. Directly handling them in the form of raw graph data is challenging. Therefore, motivated by \cite{blip2}, we aim to map the graph data into a set of fixed-size latent vectors for subsequent diffusion processes.

\noindent \textbf{Encoder}. We first utilize a trained GNN $f^s_{GNN}$ to obtain node representations $H \in \mathbb{R}^{p \times d}$. Since the number of nodes $p$ can vary for each sampled subgraph, we need a way to process these variable-sized representations. 
Inspired by the Q-Former's use in computer vision \cite{swetha2024x}—where it employs learnable query embeddings to build interactions between text and visual data—we adopt a similar strategy. Specifically, we employ $K$ learnable query tokens $Q \in \mathbb{R}^{K \times d}$ to obtain a fixed-size, compact latent representation. The Q-Former processes these tokens through a series of operations:
\begin{align}
\label{eq:q-former}
Q' &= \text{SelfAttn}(Q) \\
H_Q &= \text{CrossAttn}(Q'; H) \\
Z &= \text{MLP}(H_Q)
\end{align}
Here, \textbf{SelfAttn} is the self-attention operation \cite{vaswani2017attention} that allows the query tokens to interact with each other, capturing their internal relationships to obtatin $Q' \in \mathbb{R}^{K \times d}$. The \textbf{CrossAttn} means the cross attention operation, which enables the refined query tokens to extract a fixed-size representation, $H_Q \in \mathbb{R}^{K \times d}$. Finally, a multi-layer perceptron (MLP) is employed to obtain the final compressed representation $Z \in \mathbb{R}^{K \times d}$. This representation provides a standardized input for subsequent diffusion model operations.

\noindent \textbf{Diffusion Process.}  Given that the diffusion process is designed to reconstruct original information from corrupted data \cite{mancini2019adagraph,yang2023diffusion}, we utilize a diffusion model to capture source characteristics and subsequently employ it for construcing restoration sequences. 
We model the graph restoration process on source graph using a diffusion framework as follows:
\begin{equation}
Z \xrightarrow{} Z_0 \xrightarrow{\phi} Z_1 \xrightarrow{\phi} \cdots \xrightarrow{\phi} Z_t \xrightarrow{\psi} \hat{Z}_{t-1}  \xrightarrow{\psi} \cdots \xrightarrow{\psi} \hat{Z}_0
\end{equation}

where $\phi$ and $\psi$ denote the adding noise and removing noise process. 
Followed Equation \ref{loss:diffusion}, the training objective of diffusion is formally as:
\begin{equation}
    \mathcal{L}_{diff} = \mathbb{E}_{t , Z_0, \mathbf{\epsilon_t}} \left[ \left\| \mathbf{\epsilon}_t - \mathbf{\epsilon}_{\theta}(Z_t, t) \right\|^2 \right]
\end{equation}

After training, we can construct a sequence of restored embeddings at the embedding level, denoted as $Z_{res} = \{Z_T, Z_{T-1}, ..., Z_0\}$, where $T$ is the total restoration steps. This sequence effectively tracks the restoration process from a corrupted representation back to the original characteristics.

\noindent \textbf{Quantizer.}  
Our objective is to leverage LLMs to comprehend and effectively address the graph restoration process. However, the continuous nature of the embedding sequence $Z_{res}$ is fundamentally incompatible with the discrete, token-based generation and reasoning capabilities of LLMs. To bridge this representational gap, we incorporate a quantization strategy to discretize these continuous representations \cite{peng2021generating, razavi2019generating, lee2024leveraging}. Specifically, we maintain a learnable codebook, $P = \{p_i\}_{i=1}^M$, where $M$ is the codebook size and each $p_i \in \mathbb{R}^d$ is a learnable embedding vector.
The quantization process transforms each continuous embedding $Z_{t,i}$ (the $i$-th row vector of $Z_t$) into its nearest discrete token $s_t$. This is achieved by finding the index of the closest vector in the codebook $P$ in terms of Euclidean distance:
\begin{equation}
s_{t, i} = \underset{j \in \{1, \dots, M\}}{\text{argmin}} \, \| Z_{t, i} - p_j \|_2
\end{equation}
Through this operation, the each continuous embedding sequence $Z_t$ is transformed into a discrete token sequence, $S_t = \{s_{t, K}, \dots, s_{t, 0}\}$, which can be seamlessly processed by an LLM. 

\noindent \textbf{Decoder.} 
The decoder is responsible for reconstructing the graph from the compressed discrete tokens $S$ produced by the encoder and quantizer. Its primary goal is to invert the quantization process and encoder process while preserving the integrity of the latent information. The decoder first maps the discrete token sequence $S$ back into a continuous latent space through looking up the corresponding embeddings from the codebook $P$, yielding the de-quantized embedding  $\hat{Z} \in \mathbb{R}^{K \times d}$:
\begin{equation}
    \hat{Z} = \text{Lookup}(S; P)
\end{equation}

% Similar to encoder, the decoder employ a Q-former framework to reconstruct representation. It utilize  $\mathbf{Q}_{\text{dec}} \in \mathbb{R}^{n \times d}$, which are generated by a Multi-Layer Perceptron (MLP) from the original node features $\mathbf{X}$: $\mathbf{Q}_{dec} = \text{MLP}(\mathbf{X})$. This approach allows the decoder to leverage the original graph's characteristics to guide the reconstruction.   
Similar to the encoder, the decoder employs a Q-former framework to reconstruct the representation. It utilizes query vectors $Q_{dec} \in \mathbb{R}^{p \times d}$, which are generated through $Q_{dec} = \text{MLP}(X)$. This approach allows the decoder to leverage the original graph's characteristics to guide the reconstruction. 
Furthermore, the decoding process involves a cross-attention mechanism where the dynamic queries $Q_{dec}$ interact with the de-quantized embedding $\hat{Z}$ (serving as key and value): 
\begin{equation}
    H_{rec} = \text{CrossAttn}(Q_{dec};\hat{Z})
\end{equation}
The final node features and adjacency matrix are reconstructed as follows:

\begin{align}
    \hat{X} & = \text{MLP}(H_{rec}) \\
    \hat{A}_{ij} & = \sigma(H_{rec,i}^\top H_{rec,j})
\end{align}

where $\sigma$ is the sigmod function. This structured process ensures the decoder can accurately restore both the node features and the graph's topological structure from the compressed latent space.

\noindent \textbf{Training.} We adopt an end-to-end training approach driven by a composite loss function, ensuring that each component contributes to the final graph restoration task.
% The training objective is to minimize a weighted sum of losses from the quantizer, diffusion model, and decoder. 
% The GNN encoder's training is implicitly driven by the gradients backpropagated from these downstream tasks. 
The total loss function is defined as:
\begin{equation}
\label{equ:total_loss}
\mathcal{L}_{total} = \mathcal{L}_{diff} + \lambda_1 \mathcal{L}_{quant} + \lambda_2 \mathcal{L}_{dec}
\end{equation}
where $\lambda_1, \lambda_2$ are hyperparameters controlling the weight of each loss term.

\begin{itemize}[leftmargin=*]
\setlength{\itemsep}{0pt}
\setlength{\parsep}{0pt}
\setlength{\parskip}{0pt}
    \item \textbf{Quantization Loss ($\mathcal{L}_{\text{quant}}$):} This loss ensures the continuous embeddings from the encoder can be effectively mapped to the discrete codebook. It combines a reconstruction term and a commitment term to stabilize training:
    \begin{align}
        \mathcal{L}_{quant} = \| \text{sg}[Z] - p_s \|_2^2 + \| Z - \text{sg}[p_s] \|_2^2 \\
        \text{where} \quad p_{s,i} = \underset{c \in P}{\arg\min} \| Z_i - c \|_2^2
    \end{align}
    
    Here, $\text{sg}[\cdot]$ is the stop-gradient operator. The first term pulls the codebook vectors ($p_s$) toward the encoder outputs ($Z$), while the second term pulls the encoder outputs toward the codebook vectors, preventing the embeddings from collapsing.

    \item \textbf{Decoding Loss ($\mathcal{L}_{\text{dec}}$):} This loss guarantees the decoder can accurately reconstruct the original graph from the processed latent representation. It consists of two parts: a Binary Cross-Entropy (BCE) loss for the adjacency matrix and an L2 loss for the node features:
    \begin{equation}
        \mathcal{L}_{dec} = \text{BCE}(\hat{A}, A) + \| \hat{X} - X \|_2^2
    \end{equation}
\end{itemize}

\subsection{LLM-driven Graph Alignment}
\subsubsection{LLM Pre-training on Graph Tokens}
With trained diffusion process and quantizer model, we can transform graph restoration process into discrete token sequences that LLM can understand like language. The tranferred token symbols are viewed as new tokens needed to further learn.  
To enable the LLM to master the graph restoration process, we first extend its vocabulary by introducing $M$ new tokens. We then fine-tune the LLM to maximize the likelihood of the graph restoration process in an autoregressive manner. Formally, given a sequence of graph $\{ S_1, S_2, ..., S_T\}$ tokens obtained from the diffusion process, where each $S_i = \{s_{i,1}, s_{i,2}, ... s_{i, K}\}$, the fine-tuning objective is to minimize the negative log-likelihood as follows:
\begin{equation}
    \mathcal{L}_{LLM} = - \frac{1}{T \cdot K } \sum_{i=1}^{T} \sum_{j=1}^{K} \log P(s_{i,j} | S_{<i}, s_{i,<j})
\end{equation}

where $S_{<i}$ represents the entire sequence of tokens from all previous restoration steps ($S_1, S_2, \dots, S_{i-1}$). $s_{i,<j}$ represents the sequence of tokens from the current step $i$ that have already been generated ($s_{i,1}, s_{i,2}, \dots, s_{i,j-1}$). In this way, LLM will learn how to generate needed graph tokens to represent graph characteristics.  

\subsubsection{Post-Training for Source Alignment}

To enhance the alignment of graph tokens generated by our LLM with the implicit characteristics of the source domain, we employ reinforcement learning (RL) fine-tuning. This approach, inspired by reasoning models, guides the LLM to capture and restore the implicit properties of the source data during the graph refinement process.

% To enhance the alignment of graph tokens generated by our LLM with the implicit characteristics of the source domain, we employ reinforcement learning fine-tuning. This approach, inspired by reasoning models, guides the LLM to capture and restore the implicit properties of the source data during the graph refinement process. 

Here, we utilize Group Relative Policy Optimization (GRPO) \cite{ds-r1} for fine-tuning. Unlike standard RL algorithms that rely on a separate critic model like PPO \cite{ppo}, GRPO directly leverages a group of candidate responses to estimate the advantage. Given an input refined graph $\hat{G}^t$, GRPO first samples $g$ distinct output sequences $\{o_1, o_2, \dots, o_g\}$ from the old policy $\pi_{\text{old}}$ and computes their corresponding rewards $\{r_1, r_2, \dots, r_g\}$. The advantage $A_i$ for each output $o_i$ is then computed by normalizing its reward against the batch statistics:
\begin{equation}
    A_i = \frac{r_i - \text{mean}(\{r_1, ..., r_g\})}{\text{std}(\{r_1, ..., r_g\})}
\end{equation}
where \text{mean} and \text{std} represent the mean and standard deviation of the reward group, respectively.

The RL objective is defined to maximize the expected reward while penalizing large policy shifts, ensuring stable training:

\begin{align}
\max_{\pi_\theta} \mathbb{E}_{o \sim \pi_\theta(\hat{G}^t)} \left[ R(\hat{G}^t, o) - \beta \mathbb{D}_{\text{KL}}[\pi_\theta(o|\hat{G}^t) || \pi_{\text{old}}(o|\hat{G}^t)] \right]
\end{align}

% \begin{align}
%     \max_{\pi_\theta} & \mathbb{E}_{o \sim \pi_\theta(\hat{G}^t)} \\= & R(\hat{G}^t, o) - \beta \mathbb{D}_{\text{KL}}[\pi_\theta(o|\hat{G}^t) || \pi_{old}(o|\hat{G}^t)]
% \end{align}

where $\mathbb{D}_{\text{KL}}$ is the KL-divergence, a measure of policy divergence, and $\beta$ is a hyperparameter that controls the weight of the KL penalty. $R(\hat{G}^t, o)$ is the reward function.

We design a composite reward function to guide the LLM's refinement process, comprising two main components: an alignment reward and a confidence reward. 
The alignment reward, $R_{align}$, ensures that the generated graph's node embedding distribution statistically aligns with that of the source domain. We use Maximum Mean Discrepancy (MMD) as our metric \cite{mmd, sun2024mmd}, which measures the distance between two probability distributions in a reproducing kernel Hilbert space. The squared MMD distance, $d^2_{\text{MMD}}$, is calculated between two sets of node embeddings: one from the refined graph and one from the original source graph. Given the constraints of the TT-GDA task, we cannot directly access the source domain data. To address this, we maintain a cluster centroid matrix, $X^{s,c} \in \mathbb{R}^{C \times d}$, as a statistical representation of the source domain. This is computed by simply averaging the vectors within each class. In this manner, we obtain an approximate distribution of the source domain without explicitly accessing the source data.
% While in the TT-GDA task, we can't directly access the original source data, we maintain a cluster centroid $X^{s,c} \in \mathbb{R}^{C \times d}$ to represent the source domain's distribution by simply averaging the source embeddings.   
Let $\hat{X}^t$ be the sets of node embeddings for the refined graph with $C'$ nodes. The squared MMD distance is calculated as follows:
% The MMD (Maximum Mean Discrepancy) reward ensures that the refined graph's characteristics statistically align with those of the source domain. MMD measures the distance between two probability distributions based on their embeddings in a reproducing kernel Hilbert space. The squared MMD distance between the final refined graph's embedding representation $\hat{X}^t$ (with $n_1$ nodes) and the original source graph's embedding representation $X^s$ (with $n_2$ nodes) is calculated as follows:
%
\begin{align}
    d^2_{MMD} = & \frac{1}{C'(C'-1)} \sum_{i \neq j}^{n_1} k(\hat{X}^t_i, \hat{X}^t_j) + \frac{1}{C(C-1)} \sum_{i \neq j}^{n_2} k(X^{s,c}_i, X^{s,c}_j) \nonumber \\
    & - \frac{2}{C C'} \sum_{i=1}^{C'} \sum_{j=1}^{C} k(\hat{X}^t_i, X^{s,c}_j)
\end{align}
where $k(\cdot, \cdot)$ is a kernel function, such as the Gaussian kernel, which computes the similarity between node embeddings. To convert this distance into a reinforcement learning signal, we define the alignment reward as:
% where $k(\cdot, \cdot)$ is a kernel function, such as the Gaussian kernel, that computes the similarity between node embeddings. Further, we define the MMD reward function to convert the distance to alignment signal as below:
% \begin{equation}
%     R_{align} = e^{- \gamma d^2_{MMD}}
% \end{equation}
\begin{equation}
    R_{align} = e^{- \gamma d^2_{MMD}}
\end{equation}
where $\gamma > 0$ is a hyperparameter that controls the reward decay rate. This formulation ensures that $R_{\text{align}} \in (0, 1]$, providing a stable and well-bounded reward signal.

As a gold standard for downstream performance is unavailable, we instead use the prediction confidence of a pre-trained GNN model $f^s_{GNN}$ as a proxy for the quality of the refined graph. We define the downstream reward as the average negative entropy of the GNN's predictions on the refined graph.
The reward is calculated based on predicted probabilities of $f^s_{GNN}$ for each node in the refined graph $\hat{G}^t$:

% \begin{equation}
%     R_{\text{conf}} = -\frac{1}{n_1} \sum_{i=1}^{n_1} \sum_{c=1}^{C} p_{i,c} \log p_{i,c}
% \end{equation}
\begin{equation}
R_{conf} = -\frac{1}{C'} \sum_{i=1}^{C'} \sum_{c=1}^{C} p_{i,c} \log p_{i,c}
\end{equation}

where $p_{i,c}$ denotes the probability of node $i$ belonging to class $c$. Maximizing this reward encourages the LLM to generate graphs that lead to more confident predictions by the downstream GNN, thereby improving its performance on the target domain. Consequently, the final reward function for a generated graph is a composite of the two defined rewards:
\begin{equation}
    R_{final} = R_{align} + R_{conf}
\end{equation}
This dual-component reward guides the model to both align its output with source characteristics and improve the predictive confidence of the refined graph.  
% MMD 和 下游任务损失 作为reward

%% file: experiments.tex
\section{Experiments}
We conduct comprehensive experiments to address the following key research questions:
\begin{itemize}[leftmargin=*]
    \item RQ1: Does our proposed method achieve significant performance improvements on the node-level Test-Time Graph Domain Adaptation task?
    \item RQ2: Do the various components of our proposed approach each contribute to the final performance?
    \item RQ3: What is the impact of our approach on the quality of target graph refinement?
    % tokenization scheme capture the graph restoration process?
    % \item Does our reward design effectively align the target graph with the source domain's characteristics?
    \item RQ4: How does the proposed method's performance vary with key hyperparameters, such as the loss weight $\lambda_1, \lambda_2$, codebook size $M$ and refinement steps $T$?
\end{itemize}

\begin{table}[th]
    \centering
    \caption{Statistics of the datasets.}
    \vspace{-0.3cm}
    \resizebox{\linewidth}{!}{
    \begin{tabular}{c|cccc}
        \toprule
        Datasets & \#Nodes & \#Edges  & \#Features & \#Labels \\ 
        \midrule
        DBLPv7 & 5,484  & 4412 & 6,755 & 5\\
        ACMv9 & 9,360 & 15,602 & 6,775   & 5\\
        Citationv1 & 8,935  & 5379 & 6,755  &  5\\

        \bottomrule
    \end{tabular}
    }
    \label{tab:dataset}
    % \vspace{-0.5cm}
\end{table}

\subsection{Experiment Settings}
\textbf{Datasets.} 
For our experiments, we follow prior work \cite{udagcn, mao2024source} and utilize three widely adopted real-world network datasets from the \textbf{ArnetMiner} collection \cite{tang2008arnetminer}: ACMv9, Citationv1, and DBLPv7.
These three datasets are citation networks from distinct sources and time periods: ACMv9 (after 2010), Microsoft Academic Graph (before 2008), and DBLP (from 2004 to 2008). We represent each network as an undirected graph, where each node corresponds to a paper and an edge signifies a citation relationship between two papers. Each paper is assigned to one of five research categories: \textit{Artificial Intelligence}, \textit{Computer Vision}, \textit{Database}, \textit{Information Security}, and \textit{Networking}. For simplicity, these datasets are denoted as A, C, and D, respectively. Due to their distinct sources and collection periods, these datasets exhibit significant differences in both feature distributions and topological structures, making them ideal for evaluating domain adaptation methods. A detailed statistical overview of these datasets is provided in Table \ref{tab:dataset}.
% For our experiments, we follow prior work \cite{udagcn, mao2024source} and use three widely adopted real-world network datasets from ArnetMiner \cite{tang2008arnetminer}: ACMv9, Citationv1, and DBLPv7. These three datasets
% are citation networks from different sources: ACM (after 2010), Microsoft Academic Graph (before 2008), and DBLP (from 2004 to 2008). We represent each citation network as an undirected graph where each node denotes a paper, and an edge corresponds to a citation relationship between two papers. Each paper belongs to one of the following five categories based on its research topics: \textit{Artificial Intelligence}, \textit{Computer Vision}, \textit{Database}, \textit{Information Security}, and \textit{Networking}. These are denoted as A, C, and D for simplicity. Due to their distinct sources, these datasets exhibit different data distributions, making them ideal for evaluating domain adaptation methods. Detailed statistics are available in Table \ref{tab:dataset}. \\

\noindent \textbf{Baselines.} We mainly consider the following baseline TT-GDA methods for comparison:
\begin{itemize}[leftmargin=*]
    \item \textbf{Vanilla GNN Methods:} This category includes standard GNN models such as GCN \cite{gcn}, GSAGE \cite{GSAGE}, and GAT \cite{gat}. These models are trained on the source graph and then directly evaluated on the target graph without any adaptation, serving as a performance lower bound.
    \item \textbf{Model-Centric TT-GDA Methods:} These approaches focus on adjusting the pre-trained model's parameters to improve adaptation. We include EERM \cite{eerm}, SOGA \cite{soga}, and Matcha \cite{matcha} in this group. They typically employ different self-supervised loss functions or unique architectural designs to achieve adaptation.
    \item \textbf{Data-Centric TT-GDA Methods:} These methods, represented by GTrans \cite{gtrans} and GraphPatcher \cite{graphpatcher}, aim to modify the target graph itself to better align it with the source graph's distribution. This allows the pre-trained GNN to perform better without needing to alter its parameters.
\end{itemize}
% We mainly consider following base TT-GDA methods: (1) Vanilla GNN methods: GCN \cite{gcn}, GSAGE \cite{GSAGE} and GAT \cite{gat}. These approaches are trained on source graph and directly evaluated on target graph. (2) model-centric TT-GDA methods: EERM \cite{EERM}, SOGA \cite{soga}, Matcha \cite{matcha}. These method mainly desgin different model framewokrs or self-supervised loss to adjust model paramters for better adaptation.   (3) data-centric TT-GDA methods: GTrans \cite{gtrans} and GraphPatcher \cite{graphpatcher}. These methods are mainly adjust the graph to aligm with source graphs. 

\noindent \textbf{Metrics.} We evaluate our model's performance on the target graph using the Micro-F1 and Macro-F1 scores for classification.\\

\noindent \textbf{Implementation Details.} 
% We select GCN as the default source-trained GNN model, $f^s_{GNN}$. For a fair comparision, we set our method and all baselines's hidden size to 256. For our encoder and diffusion processes training, we randomly sample source subgraphs up to their 3-hop neighbors. The diffusion model's denoising steps, $T$, are set to 10. We use a codebook size, $C$, of 128. The loss weights $\lambda_1$ and $\lambda_2$ are set to 0.4 and 1.0, respectively. For LLM-driven Graph Alignment, we use the Llama3.1-8B as our initial pre-trained model. The learning rates for the SFT and post-training processes are $1 \times 10^{-4}$ and $2 \times 10^{-6}$, respectively.
We select the GCN as the default source-trained GNN model, denoted as $f^s_{GNN}$. For a fair comparison, we set the hidden size for both our method and all baselines to 256. For the training of our encoder and diffusion model, we randomly sample source subgraphs up to their 3-hop neighbors. The number of learnable query tokens $K$ is set to 128. The number of denoising steps for the diffusion process, $T$, is set to 10, and the codebook size, $C$, is 128. The loss weights $\lambda_1$ and $\lambda_2$ are configured as 0.4 and 1.0, respectively. In the LLM-driven graph alignment stage, we utilize the Llama3.1-8B as our base pre-trained model. The learning rates for SFT and subsequent reinforcement learning processes are set to $1 \times 10^{-4}$ and $2 \times 10^{-6}$, respectively.

% \begin{table*}[htbp]
% \centering
% \caption{The performance (\%) of all models is evaluated across six domain adaptation scenarios. The best results are \textbf{highlighted} and the second-best results are \underline{underlined}.}
% \label{tab:results}
% \resizebox{\linewidth}{!}{
% \begin{tabular}{l|cccccccccccc}
% \toprule
% \multirow{2}{*}{Methods} & \multicolumn{2}{c}{C$\Rightarrow$A} & \multicolumn{2}{c}{D$\Rightarrow$A} & \multicolumn{2}{c}{A$\Rightarrow$C} & \multicolumn{2}{c}{D$\Rightarrow$C} & \multicolumn{2}{c}{A$\Rightarrow$D} & \multicolumn{2}{c}{C$\Rightarrow$D} \\
% & Micro & Macro & Micro & Macro & Micro & Macro & Micro & Macro & Micro & Macro & Micro & Macro \\ \midrule
% $\text{GCN}$ & 65.15 & 64.11 & 58.68 & 53.35 & 72.46 & 64.34 & 69.37 & 66.10 & 65.27 & 61.80 & 66.52 & 64.24\\
% $\text{GSAGE}$ & 62.34 & 60.73 & 52.13 & 51.67 & 66.31 & 64.10 & 68.13 & 65.73 & 63.12 & 60.09 & 65.17 & 63.19 \\
% $\text{GAT}$ & 64.15 & 62.25 & 57.23 & 54.50 & 73.01 & 65.23 &  70.14 & 67.10 & 66.82 & 65.78 & 66.12 & 65.43\\
% \midrule
% $\text{EERM}$ & 71.73 & 69.87 & 68.94 & 68.73 & & & & & & & & \\
% $\text{SOGA}$ & 72.19 & 70.15  & 70.61 & 69.28 & & & & & & & & \\
% $\text{Matcha}$ & 72.49 & 71.03 & 71.48 & 69.33 & & & & & &  \\
% \midrule
% $\text{GTrans}$ & & & 69.72 & 68.37 & & & & & & & & \\
% $\text{GraphPatcher}$ & & & 70.17 & 69.09 & & & & & & & & \\

% \midrule
% \method{} & & & \textbf{72.19} & \textbf{71.07} & & & & & & & & \\
% $\text{Improv.(\%)}$ & & & & & & & & & & & & \\
% \bottomrule
% \end{tabular}
% }
% \end{table*}

\begin{table*}[htbp]
\centering
\caption{The performance (\%) of all models is evaluated across six domain adaptation scenarios. The best results are \textbf{highlighted} and the second-best results are \underline{underlined}.}
\label{tab:results}
\resizebox{\linewidth}{!}{
\begin{tabular}{l|cccccccccccc}
\toprule
\multirow{2}{*}{Methods} & \multicolumn{2}{c}{C$\Rightarrow$A} & \multicolumn{2}{c}{D$\Rightarrow$A} & \multicolumn{2}{c}{A$\Rightarrow$C} & \multicolumn{2}{c}{D$\Rightarrow$C} & \multicolumn{2}{c}{A$\Rightarrow$D} & \multicolumn{2}{c}{C$\Rightarrow$D} \\
& Micro & Macro & Micro & Macro & Micro & Macro & Micro & Macro & Micro & Macro & Micro & Macro \\ \midrule
$\text{GCN}$ & 65.15 & 64.11 & 58.68 & 53.35 & 72.46 & 64.34 & 69.37 & 66.10 & 65.27 & 61.80 & 66.52 & 64.24 \\
$\text{GSAGE}$ & 62.34 & 60.73 & 52.13 & 51.67 & 66.31 & 64.10 & 68.13 & 65.73 & 63.12 & 60.09 & 65.17 & 63.19 \\
$\text{GAT}$ & 64.15 & 62.25 & 57.23 & 54.50 & 73.01 & 65.23 & 70.14 & 67.10 & 66.82 & 65.78 & 66.12 & 65.43 \\
\midrule
$\text{EERM}$ & 69.73 & 67.87 & 68.94 & 68.73 & 74.75 & 70.20 & 74.12 & 72.42 & 68.66 & 65.37 & 67.86 & 67.28 \\
$\text{SOGA}$ & 70.19 & 68.15 & \underline{70.61} & 69.28 & 76.90 & 70.31 & 76.73 & 74.04 & 67.42 & 66.61 & 68.58 & 68.58 \\
$\text{Matcha}$ & 70.49 & 69.03 & 70.48 & \underline{69.33} & \underline{77.46} & 71.31 & 77.20 & 75.71 & 68.36 & 68.05 & 70.22 & 69.73 \\
\midrule
$\text{GTrans}$ & 71.91 & 70.40 & 69.72 & 68.37 & 77.18 & 71.22 & 76.13 & 75.93 & 69.61 & 68.37 & 70.24 & 70.07 \\
$\text{GraphPatcher}$ & \underline{72.38} & \underline{71.03} & 70.17 & 69.09 & 77.09 & \underline{72.34} & \underline{77.90} & \underline{75.94} & \underline{70.12} & \underline{69.88} & \underline{71.99} & \underline{72.82} \\
\midrule
\textbf{\method{}} & \textbf{74.59} & \textbf{73.86} & \textbf{72.19} & \textbf{71.07} & \textbf{79.58} & \textbf{74.88} & \textbf{79.49} & \textbf{77.75} & \textbf{71.64} & \textbf{71.93} & \textbf{74.37} & \textbf{74.11} \\
$\textbf{Improv.}(\text{\%})$ & +2.21 & +2.83 & +1.71 & +1.74 & +2.12 & +2.54 & +1.59 & +1.81 & +1.52 & +2.05 & +2.38 & +1.29 \\
\bottomrule
\end{tabular}
}
\end{table*}

\subsection{Main Results (RQ1)}
As shown in Table \ref{tab:results}, our approach consistently outperforms all baseline methods across every TT-GDA scenario. For instance, our method achieves an average 2.04\% improvement in the Macro-F1 score. This demonstrates that our method effectively leverages LLMs to refine graphs and imbue the target graph with latent source characteristics, leading to superior performance.
Overall, Both model-centric and data-centric TT-GDA methods generally outperform vanilla GNNs, highlighting the inherent challenge of this task and the necessity for specialized solutions. The data-centric approaches, in particular, appear to achieve better overall performance than the model-centric ones. Our method, which is a data-centric approach, significantly surpasses existing alternatives in this category. This superior performance can be attributed to several key factors: Our method combines a diffusion process with an LLM, treating the graph as a sequence of tokens. This enables a more powerful restoration process, as the LLM can generate a refined graph structure and node features that are more effectively aligned with the source domain. 
We introduced a reinforcement learning process that effectively enhances the quality of the target graph's refinement. This is achieved through specially designed alignment and confidence rewards, which provide a clear feedback signal to the LLM. This guidance encourages the LLM to make more effective changes, ultimately improving performance on the target domain.

% Our approach combines a diffusion process with an LLM, treating the graph as a sequence of tokens. This allows for a more powerful restoration process, as the LLM can generate a refined graph structure and node features that are more aligned with the source domain. By introducing rewards during the reinforcement learning process, we can further enhance the quality of the target graph's refinement. This mechanism provides a clear feedback signal to the LLM, guiding it to make more effective changes that improve performance on the target domain.

% Our approach addresses this need by transforming the TT-GDA problem into a graph restoration task, using an LLM fine-tuned with reinforcement learning to align the target graph with the source domain's characteristics. Our proposed approach can also be viewed as a kind of data-centric method. And our method significat outperform the other data-centric approach. This may be attributed our approach can combine diffusion process and LLM for token sequence to implement stronger restoration pcross. Besides, thorugh introduced rewards in reinforce learning, we can futher improve the refinement qualilty of target graph. 

\subsection{Ablation Studies (RQ2)}
To assess the contribution of each component in our method, we conduct extensive experiments comparing the following variants:
\begin{itemize}[leftmargin=*]
    \item \textbf{w/o Encoder}: This variant removes the Q-former design and instead randomly selects a fixed number of nodes as the initial input for the diffusion process.
    \item \textbf{w/o Diff}: This variant omits the diffusion process. It instead constructs the graph restoration sequences by randomly removing or adding edges.
    \item \textbf{w/o Align}: This variant removes the alignment reward, which is based on the MMD distance.
    \item \textbf{w/o Conf}: This variant removes the confidence reward, which is based on entropy calculation.
\end{itemize}

\begin{figure}[h]
    \centering
    % \subfigtopskip=1pt
    % \subfigbottomskip=1pt
    % \vspace{-0.5cm}
    \includegraphics[width=1.0\linewidth]{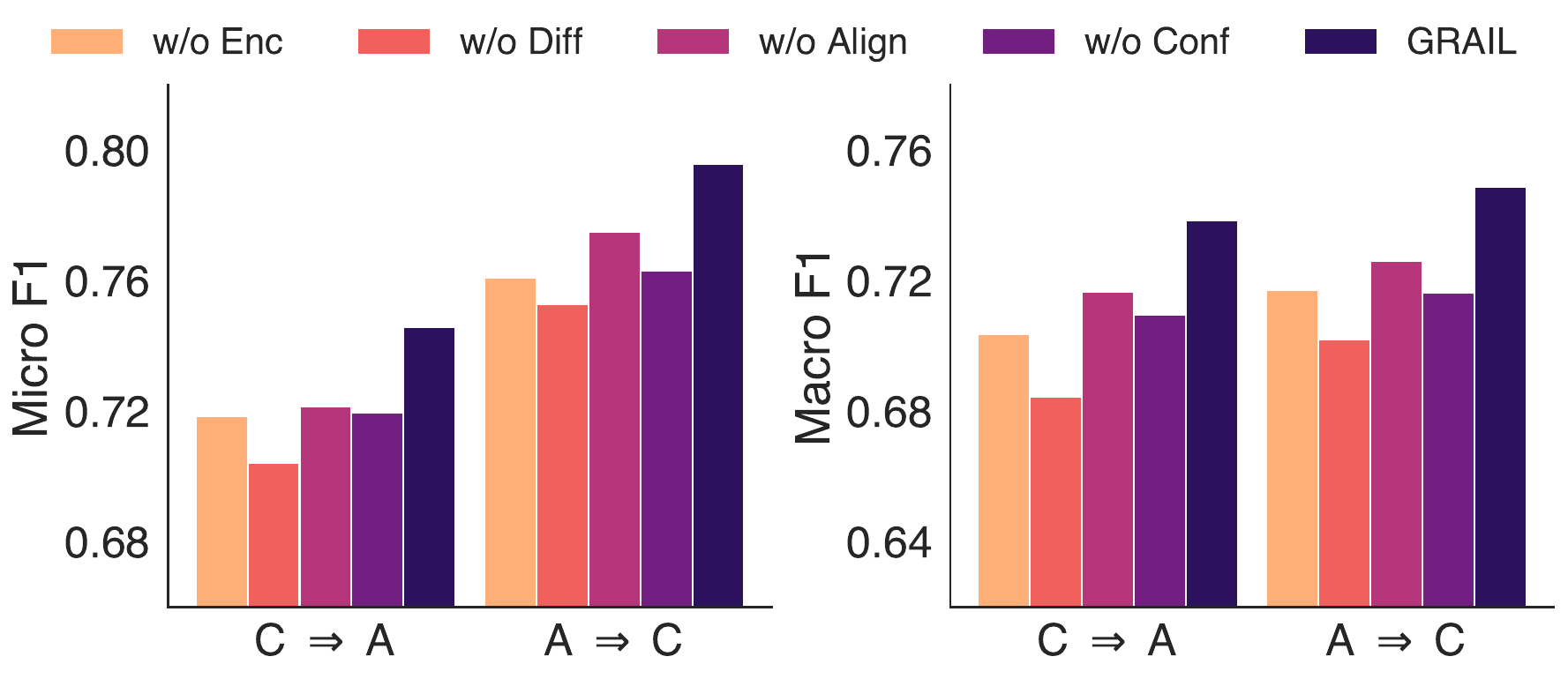}
    \caption{Ablation studies on C $\Rightarrow$ A and A $\Rightarrow$ C.}
    \label{fig:ablation}
\end{figure}

As illustrated in Figure \ref{fig:ablation}, the performance degradation of the \textbf{w/o Encoder} variant shows that our Q-former-based encoder effectively encodes a variable number of nodes into a fixed, compact latent representation for subsequent training.
The performance drop observed in the \textbf{w/o Diff} variant demonstrates that the diffusion process is effective at modeling the graph restoration process, which is crucial for recovering source characteristics. This forms a solid foundation for the subsequent LLM fine-tuning and post-training steps.
Furthermore, GRAIL's superior performance over the \textbf{w/o Align} and \textbf{w/o Conf} variants highlights the effectiveness of our designed rewards in the reinforcement learning phase. The alignment reward ensures the refined target graph maintains statistical properties aligned with the source domain, while the confidence reward prevents the refinement process from sacrificing quality for the sake of simply increasing alignment, thereby ensuring the integrity of the final graph. 

% replace q-former & quantifier 
% replace grpo 

\begin{figure}[t]
    \centering
    \subfigure[Before: Class Distribution]{
    \includegraphics[width=0.47\linewidth]{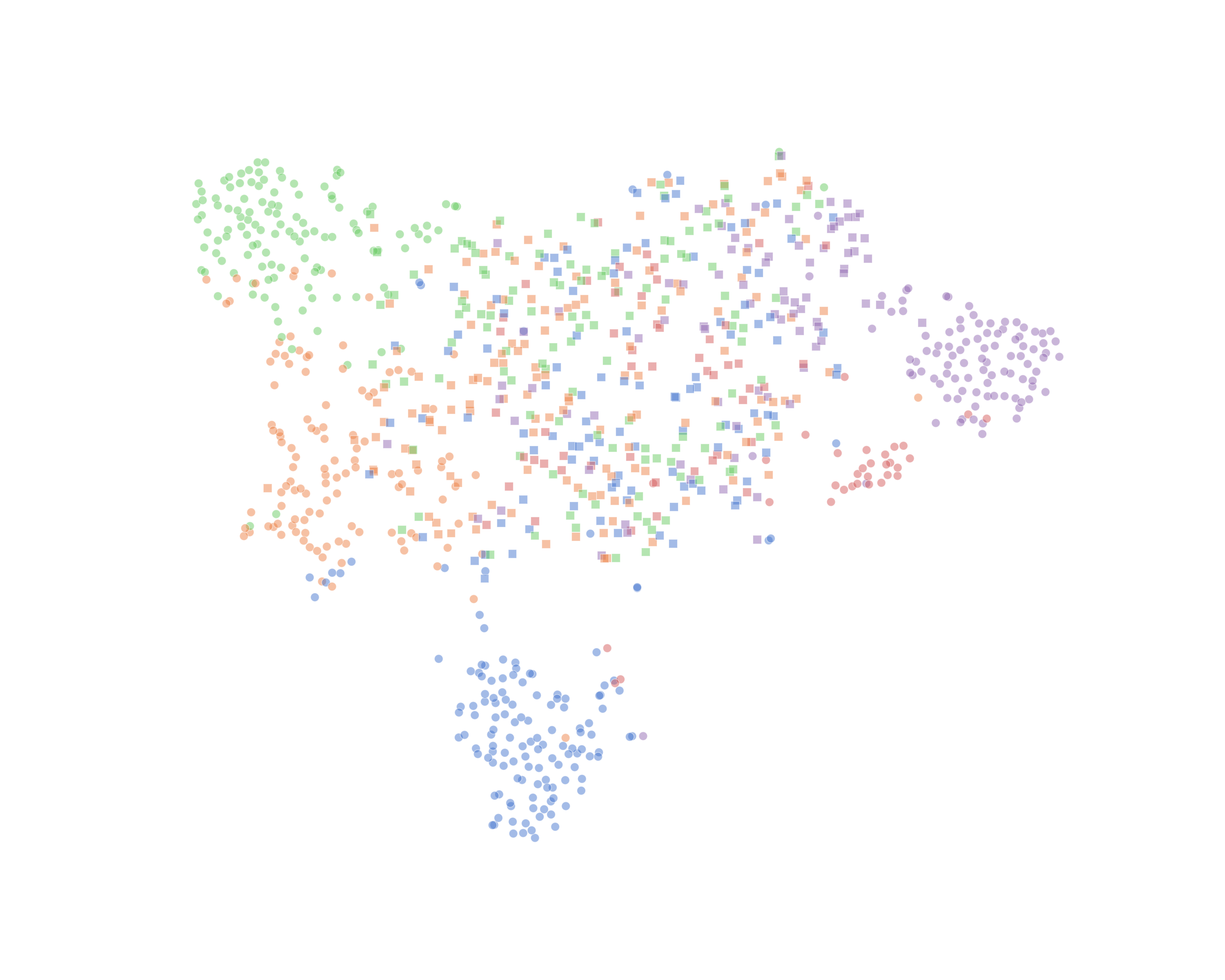}
    \label{fig:vis:before_class}
    }
    \subfigure[After: Class Distribution]{
    \includegraphics[width=0.47\linewidth]{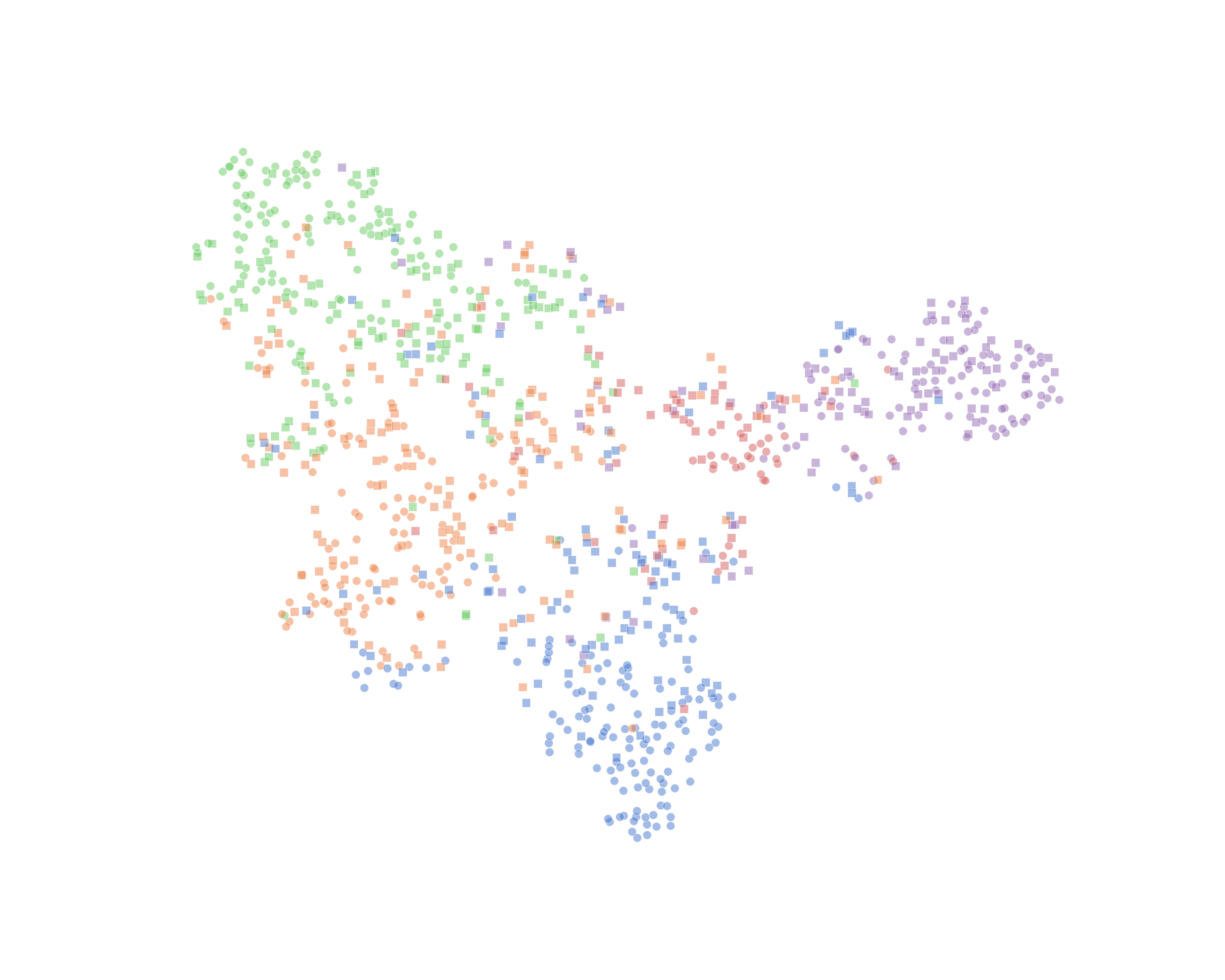}
    \label{fig:vis:after_class}
    }
    \\ % forces a new line
    \subfigure[Before: Domain Distribution]{
    \includegraphics[width=0.47\linewidth]{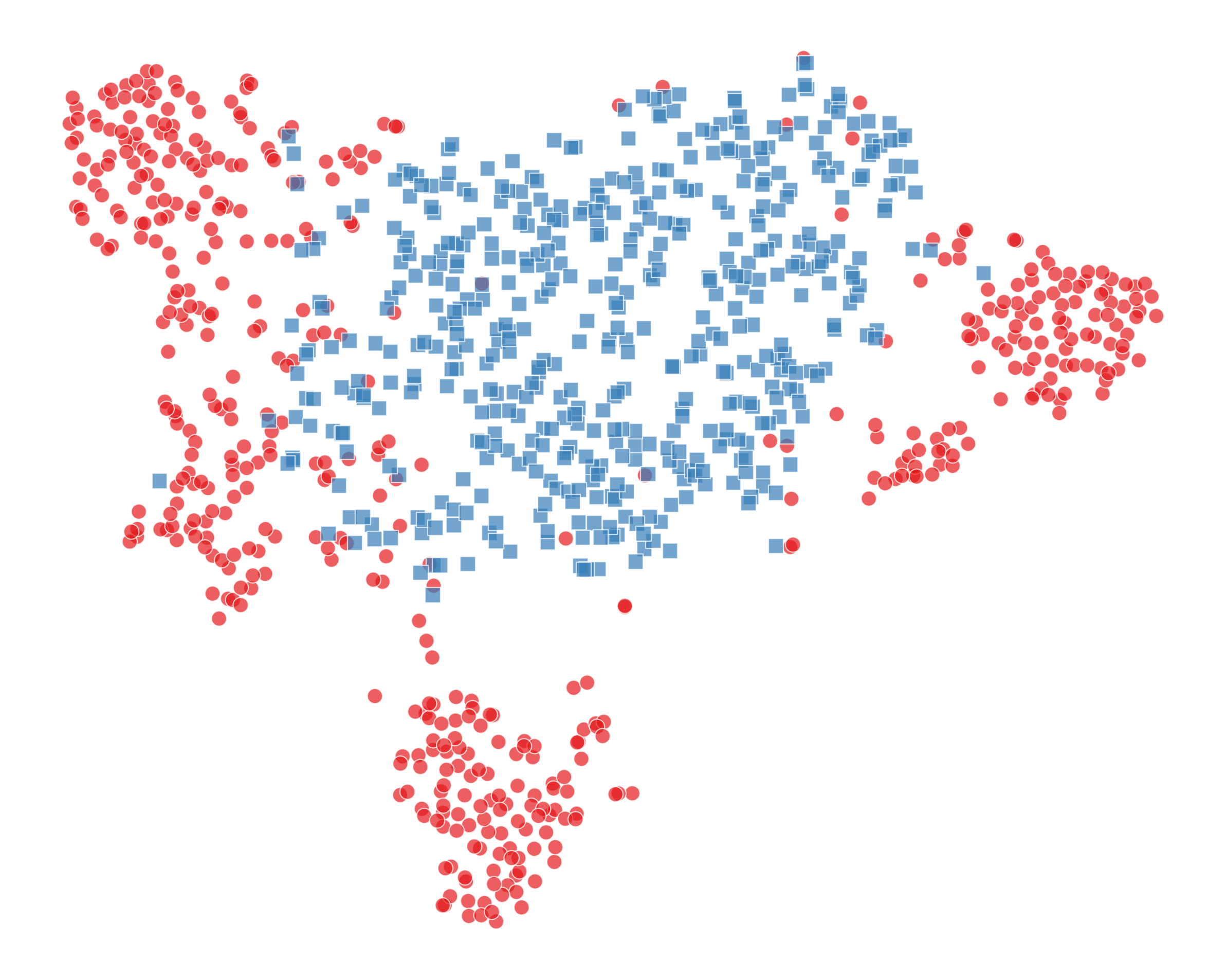}
    \label{fig:vis:before_domain}
    }
    \subfigure[After: Domain Distribution]{
    \includegraphics[width=0.47\linewidth]{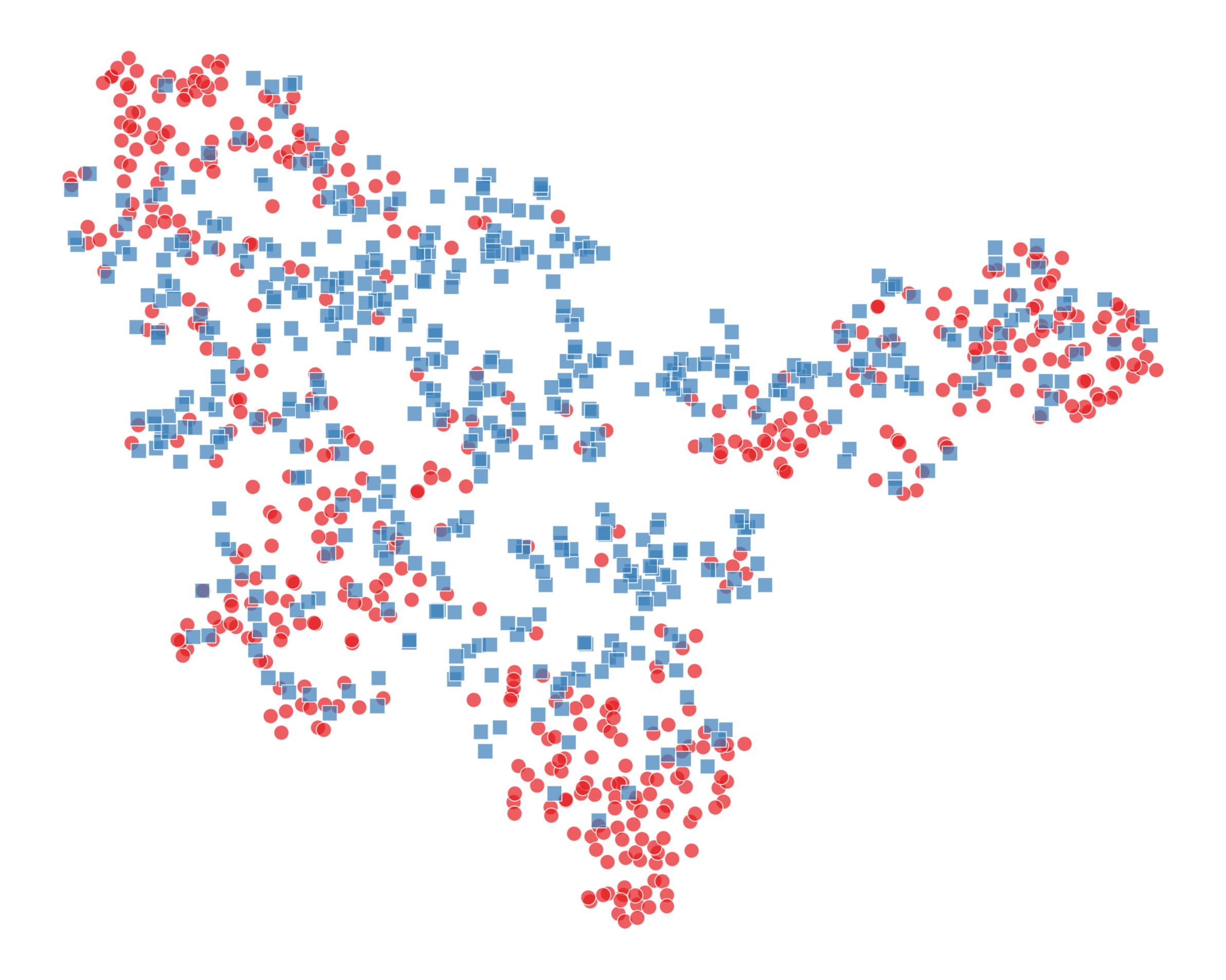}
    \label{fig:vis:after_domain}
    }
    \vspace{-1em}
    \caption{Visualization of node embeddings on the C $\Rightarrow$ A domain adaptation task. (a) and (b) show the node distributions colored by their \textit{class labels} before and after refining target subgraph, respectively. (c) and (d) show the node distributions colored by their \textit{domain type} (source or target) before and after refining.}
    \vspace{-1em}
    \label{fig:vis}
\end{figure}

\subsection{Effectiveness of Refined Target Graph (RQ3)}
% 训练前后embedding 分布等
% We further compare the representation distributions overlap of source graph with origin target node embedding and node representation from refined target graph. 
% To validate our propsed appraoch indeed contribute to the 
We further compare the representation distribution of the original target graph and the adapted target subgraph. We employ t-SNE \cite{t-sne} to project the learned latent representations into a 2-dimensional space, as shown in Figure \ref{fig:vis}. First, we use colors to represent the class labels and visualize the embedding distributions before and after adaptation in Figures \ref{fig:vis}(a) and (b). We observe that the distribution in (b) is more distinct and well-separated, with a notably better clustering of points in the central region compared to the mixed distribution in (a). This demonstrates that our graph refinement process effectively improves the source model's performance on the target graph by enhancing class separability. 

Furthermore, in Figures \ref{fig:vis}(c) and (d), we use color to differentiate the domains (blue for source and red for target). Through these visualizations, we observe that the adapted target nodes are better aligned with the source nodes in (d), forming a more cohesive cluster. In contrast, as shown in (c), the initial target nodes exhibit a more discrete and separate distribution from the source nodes. This successful alignment can be attributed to our introduced diffusion process and the alignment reward design, which together enable the generation of a refined target graph with latent source characteristics.

% We further compare the representation distribution difference from origin target graph and adapted target subgraph, respectively. We employ t-SNE \cite{t-sne} to project learned latent representations into 2-dim as showns in Figure \ref{fig:vis}. We first use colors to represent the class label and demonstrate embedding distribution befeore adaptation and after adaptation in \ref{fig:vis} (a) and (b).  We canobserve the distribution in (b) demonstrate more separated and large data in (a) are mixed espically in center region. This demonstrate our graph refine process imporve the source model's performance on target graph. Furthermore, we demonstrate the distrribution of source nodes and target nodes using color (blude for source domain and red for target domain )to differentiate the domain in Figure(c) and (d). Through the experinents, the adapted tragte nodes are alignmend better with source nodes. Rather, initial target nodes demonstraet more dicrete with source nodes. This may be attributed to our introduced diffsuion process and alignment reward's design to obation target graph with source characterstics. 

% \subsection{Case Study}
% demonstrate the node tokenzie process 

% \subsection{Key Motivation}
% encode 方案有效性，查看对结果的影响
% 强化学习效果的展示，visualization方面，对统计指标的验证
\subsection{Hyperparameters Analysis (RQ4)}
% $\lambda_1$, $\lambda_2$
% Denosing steps $T$
% Codebooksize $C$

\begin{table}[htbp]
    \centering
    \caption{Impact of codebook size $M$.}
    \begin{tabular}{@{}lc|ccccc@{}}
    \toprule
    \multicolumn{2}{c|}{\textbf{$M$}} &  32 & 64 & 128 & 256 & 512  \\ \midrule
    \multirow{2}{*}{C $\Rightarrow$ A} & Micro & 72.1 & 73.22 & 74.59 & 74.12 & 73.01  \\  
    \cline{2-7}
     & Macro & 71.12 & 73.15 & 73.86 & 73.60 & 73.17  \\ 
     \midrule
    \multirow{2}{*}{A $\Rightarrow$ C} & Micro & 78.83 & 79.45 & 79.85 & 79.90 & 79.12 \\ 
    \cline{2-7}
     & Macro & 74.17 & 74.72 & 74.88 & 74.60 & 73.92  \\ 
     \bottomrule
    \end{tabular}
    \label{table:codebook}
\end{table}

\begin{table}[htbp]
    \centering
    \caption{Impact of restoration steps $T$.}
    \label{tab:step}
    \begin{tabular}{@{}lc|ccccc@{}}
    \toprule
    \multicolumn{2}{c|}{\textbf{$T$}} &  2 & 5 & 10 & 20 & 30  \\ \midrule
    \multirow{2}{*}{C $\Rightarrow$ A} & Micro & 72.1 & 74.59 & 73.19 & 72.12 & 71.81  \\  
    \cline{2-7}
     & Macro & 71.12 & 73.86 & 73.86 & 72.60 & 70.17  \\ 
     \midrule
    \multirow{2}{*}{A $\Rightarrow$ C} & Micro & 78.83 & 79.85 & 77.90 & 75.90 & 73.12 \\ 
    \cline{2-7}
     & Macro & 73.17 & 74.88 & 74.10 & 73.60 & 72.17  \\ 
     \bottomrule
    \end{tabular}
\end{table}

To further analyze the effectiveness and robustness of our approach, we conducted a thorough investigation into the impact of several key hyperparameters. We investigate the influence of the weight parameters $\lambda_1$ and $\lambda_2$ (defined in Equation \ref{equ:total_loss}) on our model's performance. The results of this analysis are presented in Figure \ref{fig:weight}. 
The weight $\lambda_1$ controls the importance of the quantization loss. As demonstrated by the results, the model's performance first improves as $\lambda_1$ increases and then begins to decrease slowly. This suggests that the quantization loss is crucial for ensuring the discrete representations retain sufficient information for the LLM's predictive task. However, an overemphasis on this loss can diminish the effect of the primary diffusion loss, leading to a performance drop. Besides, the performance for $\lambda_2$ remains relatively stable across the range of $[0, 1]$. Given this stability, we set $\lambda_2 = 1$ as our default for simplicity, ensuring a consistent contribution from this term.
For the codebook size $M$, a larger codebook can capture more intricate details but also increases the number of learnable parameters, placing a greater burden on the LLM during inference. As shown in Table \ref{table:codebook}, a codebook size of $M = 128$ strikes an optimal balance. 
We also analyze the impact of the number of restoration steps $T$, on performance, as shown in Table \ref{tab:step}. A maximum number of steps was set to prevent unbounded generation. Experiments show that a relatively small number of steps, such as $T = 5$, is sufficient. Performance begins to decrease beyond this point. This may be attributed to error accumulation, as the LLM's sequential reasoning over more steps can degrade the quality of the final graph restoration.

% We further investigate the impact of weight parameters $\lambda_1$ and $\lambda_2$ in Equation \ref{eq:f} ans demonstrate the experiments in Figure \ref{fig:params}. $\lambda_2$ control the importance of quantizer loss \ref{equ:total_loss}. Through the results, we can conclude the performance first improve and then start to decrease slowly. This is meanly becaure the quantation loss wieght ensure discretation process maintain enough information for LLM prediction. However, when weight overempasis, it will decrease the main diffusion loss's effect. Similar condictions occurs in Figure \ref{fig:}, the perforamnce change on range[ 0-1] demonstrate relative stable. This demonstraet our degin has robustness. 
% For Codebooksize $C$, increasing it will bring more inforation, however, two much bigger will introduce too much learnable parammters and improve LLM symbol inference burden, from Tabel [], we choose $C = 128$ achiver tradeoff. 
% For denosing steps $T$, we usually rely model to output <EOS> special token to completed restroration process. Besides, we set maximum steps $T$ to advoid unlimited generation. In practicaes, $T = 5$ is enough, and then performance begin to decrease, this may be from steps will cause LLM conduct more resonabilits and cause error accumularion.   

%% file: main.bbl

\begin{thebibliography}{47}

%%% the Web address.

\ifx \showCODEN    \undefined \def \showCODEN     #1{\unskip}     \fi
\ifx \showISBNx    \undefined \def \showISBNx     #1{\unskip}     \fi
\ifx \showISBNxiii \undefined \def \showISBNxiii  #1{\unskip}     \fi
\ifx \showISSN     \undefined \def \showISSN      #1{\unskip}     \fi
\ifx \showLCCN     \undefined \def \showLCCN      #1{\unskip}     \fi
\ifx \shownote     \undefined \def \shownote      #1{#1}          \fi
\ifx \showarticletitle \undefined \def \showarticletitle #1{#1}   \fi
\ifx \showURL      \undefined \def \showURL       {\relax}        \fi
% invisible to TeX
\providecommand\bibfield[2]{#2}
\providecommand\bibinfo[2]{#2}
\providecommand\natexlab[1]{#1}
\providecommand\showeprint[2][]{arXiv:#2}

\bibitem[Achiam et~al\mbox{.}(2023)]%
        {gpt4}
\bibfield{author}{\bibinfo{person}{Josh Achiam}, \bibinfo{person}{Steven Adler}, \bibinfo{person}{Sandhini Agarwal}, \bibinfo{person}{Lama Ahmad}, \bibinfo{person}{Ilge Akkaya}, \bibinfo{person}{Florencia~Leoni Aleman}, \bibinfo{person}{Diogo Almeida}, \bibinfo{person}{Janko Altenschmidt}, \bibinfo{person}{Sam Altman}, \bibinfo{person}{Shyamal Anadkat}, {et~al\mbox{.}}} \bibinfo{year}{2023}\natexlab{}.
\newblock \showarticletitle{Gpt-4 technical report}.
\newblock \bibinfo{journal}{\emph{arXiv preprint arXiv:2303.08774}} (\bibinfo{year}{2023}).
\newblock


\bibitem[Bao et~al\mbox{.}(2024)]%
        {matcha}
\bibfield{author}{\bibinfo{person}{Wenxuan Bao}, \bibinfo{person}{Zhichen Zeng}, \bibinfo{person}{Zhining Liu}, \bibinfo{person}{Hanghang Tong}, {and} \bibinfo{person}{Jingrui He}.} \bibinfo{year}{2024}\natexlab{}.
\newblock \showarticletitle{Matcha: Mitigating Graph Structure Shifts with Test-Time Adaptation}.
\newblock \bibinfo{journal}{\emph{arXiv preprint arXiv:2410.06976}} (\bibinfo{year}{2024}).
\newblock


\bibitem[Dai et~al\mbox{.}(2022)]%
        {adagcn}
\bibfield{author}{\bibinfo{person}{Quanyu Dai}, \bibinfo{person}{Xiao-Ming Wu}, \bibinfo{person}{Jiaren Xiao}, \bibinfo{person}{Xiao Shen}, {and} \bibinfo{person}{Dan Wang}.} \bibinfo{year}{2022}\natexlab{}.
\newblock \showarticletitle{Graph transfer learning via adversarial domain adaptation with graph convolution}.
\newblock \bibinfo{journal}{\emph{IEEE Transactions on Knowledge and Data Engineering}} \bibinfo{volume}{35}, \bibinfo{number}{5} (\bibinfo{year}{2022}), \bibinfo{pages}{4908--4922}.
\newblock


\bibitem[Davies and Ajmeri(2022)]%
        {davies2022realistic}
\bibfield{author}{\bibinfo{person}{Alex Davies} {and} \bibinfo{person}{Nirav Ajmeri}.} \bibinfo{year}{2022}\natexlab{}.
\newblock \showarticletitle{Realistic synthetic social networks with graph neural networks}.
\newblock \bibinfo{journal}{\emph{arXiv preprint arXiv:2212.07843}} (\bibinfo{year}{2022}).
\newblock


\bibitem[Fan et~al\mbox{.}(2019)]%
        {fan2019graph}
\bibfield{author}{\bibinfo{person}{Wenqi Fan}, \bibinfo{person}{Yao Ma}, \bibinfo{person}{Qing Li}, \bibinfo{person}{Yuan He}, \bibinfo{person}{Eric Zhao}, \bibinfo{person}{Jiliang Tang}, {and} \bibinfo{person}{Dawei Yin}.} \bibinfo{year}{2019}\natexlab{}.
\newblock \showarticletitle{Graph neural networks for social recommendation}. In \bibinfo{booktitle}{\emph{The world wide web conference}}. \bibinfo{pages}{417--426}.
\newblock


\bibitem[Fatemi et~al\mbox{.}(2023)]%
        {fatemi2023talk}
\bibfield{author}{\bibinfo{person}{Bahare Fatemi}, \bibinfo{person}{Jonathan Halcrow}, {and} \bibinfo{person}{Bryan Perozzi}.} \bibinfo{year}{2023}\natexlab{}.
\newblock \showarticletitle{Talk like a graph: Encoding graphs for large language models}.
\newblock \bibinfo{journal}{\emph{arXiv preprint arXiv:2310.04560}} (\bibinfo{year}{2023}).
\newblock


\bibitem[Gretton et~al\mbox{.}(2012)]%
        {mmd}
\bibfield{author}{\bibinfo{person}{Arthur Gretton}, \bibinfo{person}{Karsten~M Borgwardt}, \bibinfo{person}{Malte~J Rasch}, \bibinfo{person}{Bernhard Sch{\"o}lkopf}, {and} \bibinfo{person}{Alexander Smola}.} \bibinfo{year}{2012}\natexlab{}.
\newblock \showarticletitle{A kernel two-sample test}.
\newblock \bibinfo{journal}{\emph{The Journal of Machine Learning Research}} \bibinfo{volume}{13}, \bibinfo{number}{1} (\bibinfo{year}{2012}), \bibinfo{pages}{723--773}.
\newblock


\bibitem[Guo et~al\mbox{.}(2025)]%
        {ds-r1}
\bibfield{author}{\bibinfo{person}{Daya Guo}, \bibinfo{person}{Dejian Yang}, \bibinfo{person}{Haowei Zhang}, \bibinfo{person}{Junxiao Song}, \bibinfo{person}{Ruoyu Zhang}, \bibinfo{person}{Runxin Xu}, \bibinfo{person}{Qihao Zhu}, \bibinfo{person}{Shirong Ma}, \bibinfo{person}{Peiyi Wang}, \bibinfo{person}{Xiao Bi}, {et~al\mbox{.}}} \bibinfo{year}{2025}\natexlab{}.
\newblock \showarticletitle{Deepseek-r1: Incentivizing reasoning capability in llms via reinforcement learning}.
\newblock \bibinfo{journal}{\emph{arXiv preprint arXiv:2501.12948}} (\bibinfo{year}{2025}).
\newblock


\bibitem[Hamilton et~al\mbox{.}(2017)]%
        {GSAGE}
\bibfield{author}{\bibinfo{person}{Will Hamilton}, \bibinfo{person}{Zhitao Ying}, {and} \bibinfo{person}{Jure Leskovec}.} \bibinfo{year}{2017}\natexlab{}.
\newblock \showarticletitle{Inductive representation learning on large graphs}.
\newblock \bibinfo{journal}{\emph{Advances in neural information processing systems}}  \bibinfo{volume}{30} (\bibinfo{year}{2017}).
\newblock


\bibitem[Ho et~al\mbox{.}(2020)]%
        {ho2020denoising}
\bibfield{author}{\bibinfo{person}{Jonathan Ho}, \bibinfo{person}{Ajay Jain}, {and} \bibinfo{person}{Pieter Abbeel}.} \bibinfo{year}{2020}\natexlab{}.
\newblock \showarticletitle{Denoising diffusion probabilistic models}.
\newblock \bibinfo{journal}{\emph{Advances in neural information processing systems}}  \bibinfo{volume}{33} (\bibinfo{year}{2020}), \bibinfo{pages}{6840--6851}.
\newblock


\bibitem[Jin et~al\mbox{.}(2022)]%
        {gtrans}
\bibfield{author}{\bibinfo{person}{Wei Jin}, \bibinfo{person}{Tong Zhao}, \bibinfo{person}{Jiayuan Ding}, \bibinfo{person}{Yozen Liu}, \bibinfo{person}{Jiliang Tang}, {and} \bibinfo{person}{Neil Shah}.} \bibinfo{year}{2022}\natexlab{}.
\newblock \showarticletitle{Empowering graph representation learning with test-time graph transformation}.
\newblock \bibinfo{journal}{\emph{arXiv preprint arXiv:2210.03561}} (\bibinfo{year}{2022}).
\newblock


\bibitem[Ju et~al\mbox{.}(2023)]%
        {graphpatcher}
\bibfield{author}{\bibinfo{person}{Mingxuan Ju}, \bibinfo{person}{Tong Zhao}, \bibinfo{person}{Wenhao Yu}, \bibinfo{person}{Neil Shah}, {and} \bibinfo{person}{Yanfang Ye}.} \bibinfo{year}{2023}\natexlab{}.
\newblock \showarticletitle{Graphpatcher: mitigating degree bias for graph neural networks via test-time augmentation}.
\newblock \bibinfo{journal}{\emph{Advances in Neural Information Processing Systems}}  \bibinfo{volume}{36} (\bibinfo{year}{2023}), \bibinfo{pages}{55785--55801}.
\newblock


\bibitem[Kipf and Welling(2016)]%
        {gcn}
\bibfield{author}{\bibinfo{person}{Thomas~N Kipf} {and} \bibinfo{person}{Max Welling}.} \bibinfo{year}{2016}\natexlab{}.
\newblock \showarticletitle{Semi-supervised classification with graph convolutional networks}.
\newblock \bibinfo{journal}{\emph{arXiv preprint arXiv:1609.02907}} (\bibinfo{year}{2016}).
\newblock


\bibitem[Lee et~al\mbox{.}(2024)]%
        {lee2024leveraging}
\bibfield{author}{\bibinfo{person}{Yoonhyung Lee}, \bibinfo{person}{Younhyung Chae}, {and} \bibinfo{person}{Kyomin Jung}.} \bibinfo{year}{2024}\natexlab{}.
\newblock \showarticletitle{Leveraging VQ-VAE tokenization for autoregressive modeling of medical time series}.
\newblock \bibinfo{journal}{\emph{Artificial Intelligence in Medicine}}  \bibinfo{volume}{154} (\bibinfo{year}{2024}), \bibinfo{pages}{102925}.
\newblock


\bibitem[Li et~al\mbox{.}(2023)]%
        {blip2}
\bibfield{author}{\bibinfo{person}{Junnan Li}, \bibinfo{person}{Dongxu Li}, \bibinfo{person}{Silvio Savarese}, {and} \bibinfo{person}{Steven Hoi}.} \bibinfo{year}{2023}\natexlab{}.
\newblock \showarticletitle{Blip-2: Bootstrapping language-image pre-training with frozen image encoders and large language models}. In \bibinfo{booktitle}{\emph{International conference on machine learning}}. PMLR, \bibinfo{pages}{19730--19742}.
\newblock


\bibitem[Li et~al\mbox{.}(2025)]%
        {li2025system}
\bibfield{author}{\bibinfo{person}{Zhong-Zhi Li}, \bibinfo{person}{Duzhen Zhang}, \bibinfo{person}{Ming-Liang Zhang}, \bibinfo{person}{Jiaxin Zhang}, \bibinfo{person}{Zengyan Liu}, \bibinfo{person}{Yuxuan Yao}, \bibinfo{person}{Haotian Xu}, \bibinfo{person}{Junhao Zheng}, \bibinfo{person}{Pei-Jie Wang}, \bibinfo{person}{Xiuyi Chen}, {et~al\mbox{.}}} \bibinfo{year}{2025}\natexlab{}.
\newblock \showarticletitle{From system 1 to system 2: A survey of reasoning large language models}.
\newblock \bibinfo{journal}{\emph{arXiv preprint arXiv:2502.17419}} (\bibinfo{year}{2025}).
\newblock


\bibitem[Liang et~al\mbox{.}(2025)]%
        {liang2025comprehensive}
\bibfield{author}{\bibinfo{person}{Jian Liang}, \bibinfo{person}{Ran He}, {and} \bibinfo{person}{Tieniu Tan}.} \bibinfo{year}{2025}\natexlab{}.
\newblock \showarticletitle{A comprehensive survey on test-time adaptation under distribution shifts}.
\newblock \bibinfo{journal}{\emph{International Journal of Computer Vision}} \bibinfo{volume}{133}, \bibinfo{number}{1} (\bibinfo{year}{2025}), \bibinfo{pages}{31--64}.
\newblock


\bibitem[Liu and Ding(2024)]%
        {liu2024beyond}
\bibfield{author}{\bibinfo{person}{Shuhan Liu} {and} \bibinfo{person}{Kaize Ding}.} \bibinfo{year}{2024}\natexlab{}.
\newblock \showarticletitle{Beyond Generalization: A Survey of Out-Of-Distribution Adaptation on Graphs}.
\newblock \bibinfo{journal}{\emph{arXiv preprint arXiv:2402.11153}} (\bibinfo{year}{2024}).
\newblock


\bibitem[Liu et~al\mbox{.}(2024)]%
        {sora}
\bibfield{author}{\bibinfo{person}{Yixin Liu}, \bibinfo{person}{Kai Zhang}, \bibinfo{person}{Yuan Li}, \bibinfo{person}{Zhiling Yan}, \bibinfo{person}{Chujie Gao}, \bibinfo{person}{Ruoxi Chen}, \bibinfo{person}{Zhengqing Yuan}, \bibinfo{person}{Yue Huang}, \bibinfo{person}{Hanchi Sun}, \bibinfo{person}{Jianfeng Gao}, {et~al\mbox{.}}} \bibinfo{year}{2024}\natexlab{}.
\newblock \showarticletitle{Sora: A review on background, technology, limitations, and opportunities of large vision models}.
\newblock \bibinfo{journal}{\emph{arXiv preprint arXiv:2402.17177}} (\bibinfo{year}{2024}).
\newblock


\bibitem[Ma(2024)]%
        {ma2024improved}
\bibfield{author}{\bibinfo{person}{Jing Ma}.} \bibinfo{year}{2024}\natexlab{}.
\newblock \showarticletitle{Improved self-training for test-time adaptation}. In \bibinfo{booktitle}{\emph{Proceedings of the IEEE/CVF Conference on Computer Vision and Pattern Recognition}}. \bibinfo{pages}{23701--23710}.
\newblock


\bibitem[Mancini et~al\mbox{.}(2019)]%
        {mancini2019adagraph}
\bibfield{author}{\bibinfo{person}{Massimiliano Mancini}, \bibinfo{person}{Samuel~Rota Bulo}, \bibinfo{person}{Barbara Caputo}, {and} \bibinfo{person}{Elisa Ricci}.} \bibinfo{year}{2019}\natexlab{}.
\newblock \showarticletitle{Adagraph: Unifying predictive and continuous domain adaptation through graphs}. In \bibinfo{booktitle}{\emph{Proceedings of the IEEE/CVF Conference on Computer Vision and Pattern Recognition}}. \bibinfo{pages}{6568--6577}.
\newblock


\bibitem[Mao et~al\mbox{.}(2024a)]%
        {soga}
\bibfield{author}{\bibinfo{person}{Haitao Mao}, \bibinfo{person}{Lun Du}, \bibinfo{person}{Yujia Zheng}, \bibinfo{person}{Qiang Fu}, \bibinfo{person}{Zelin Li}, \bibinfo{person}{Xu Chen}, \bibinfo{person}{Shi Han}, {and} \bibinfo{person}{Dongmei Zhang}.} \bibinfo{year}{2024}\natexlab{a}.
\newblock \showarticletitle{Source free graph unsupervised domain adaptation}. In \bibinfo{booktitle}{\emph{Proceedings of the 17th ACM International Conference on Web Search and Data Mining}}. \bibinfo{pages}{520--528}.
\newblock


\bibitem[Mao et~al\mbox{.}(2024b)]%
        {mao2024source}
\bibfield{author}{\bibinfo{person}{Haitao Mao}, \bibinfo{person}{Lun Du}, \bibinfo{person}{Yujia Zheng}, \bibinfo{person}{Qiang Fu}, \bibinfo{person}{Zelin Li}, \bibinfo{person}{Xu Chen}, \bibinfo{person}{Shi Han}, {and} \bibinfo{person}{Dongmei Zhang}.} \bibinfo{year}{2024}\natexlab{b}.
\newblock \showarticletitle{Source free graph unsupervised domain adaptation}. In \bibinfo{booktitle}{\emph{Proceedings of the 17th ACM International Conference on Web Search and Data Mining}}. \bibinfo{pages}{520--528}.
\newblock


\bibitem[Nichol and Dhariwal(2021)]%
        {nichol2021improved}
\bibfield{author}{\bibinfo{person}{Alexander~Quinn Nichol} {and} \bibinfo{person}{Prafulla Dhariwal}.} \bibinfo{year}{2021}\natexlab{}.
\newblock \showarticletitle{Improved denoising diffusion probabilistic models}. In \bibinfo{booktitle}{\emph{International conference on machine learning}}. PMLR, \bibinfo{pages}{8162--8171}.
\newblock


\bibitem[{\"O}zbey et~al\mbox{.}(2023)]%
        {ozbey2023unsupervised}
\bibfield{author}{\bibinfo{person}{Muzaffer {\"O}zbey}, \bibinfo{person}{Onat Dalmaz}, \bibinfo{person}{Salman~UH Dar}, \bibinfo{person}{Hasan~A Bedel}, \bibinfo{person}{{\c{S}}aban {\"O}zturk}, \bibinfo{person}{Alper G{\"u}ng{\"o}r}, {and} \bibinfo{person}{Tolga Cukur}.} \bibinfo{year}{2023}\natexlab{}.
\newblock \showarticletitle{Unsupervised medical image translation with adversarial diffusion models}.
\newblock \bibinfo{journal}{\emph{IEEE Transactions on Medical Imaging}} \bibinfo{volume}{42}, \bibinfo{number}{12} (\bibinfo{year}{2023}), \bibinfo{pages}{3524--3539}.
\newblock


\bibitem[Peng et~al\mbox{.}(2021)]%
        {peng2021generating}
\bibfield{author}{\bibinfo{person}{Jialun Peng}, \bibinfo{person}{Dong Liu}, \bibinfo{person}{Songcen Xu}, {and} \bibinfo{person}{Houqiang Li}.} \bibinfo{year}{2021}\natexlab{}.
\newblock \showarticletitle{Generating diverse structure for image inpainting with hierarchical VQ-VAE}. In \bibinfo{booktitle}{\emph{Proceedings of the IEEE/CVF conference on computer vision and pattern recognition}}. \bibinfo{pages}{10775--10784}.
\newblock


\bibitem[Razavi et~al\mbox{.}(2019)]%
        {razavi2019generating}
\bibfield{author}{\bibinfo{person}{Ali Razavi}, \bibinfo{person}{Aaron Van~den Oord}, {and} \bibinfo{person}{Oriol Vinyals}.} \bibinfo{year}{2019}\natexlab{}.
\newblock \showarticletitle{Generating diverse high-fidelity images with vq-vae-2}.
\newblock \bibinfo{journal}{\emph{Advances in neural information processing systems}}  \bibinfo{volume}{32} (\bibinfo{year}{2019}).
\newblock


\bibitem[Ren et~al\mbox{.}(2024)]%
        {ren2024survey}
\bibfield{author}{\bibinfo{person}{Xubin Ren}, \bibinfo{person}{Jiabin Tang}, \bibinfo{person}{Dawei Yin}, \bibinfo{person}{Nitesh Chawla}, {and} \bibinfo{person}{Chao Huang}.} \bibinfo{year}{2024}\natexlab{}.
\newblock \showarticletitle{A survey of large language models for graphs}. In \bibinfo{booktitle}{\emph{Proceedings of the 30th ACM SIGKDD Conference on Knowledge Discovery and Data Mining}}. \bibinfo{pages}{6616--6626}.
\newblock


\bibitem[Schulman et~al\mbox{.}(2017)]%
        {ppo}
\bibfield{author}{\bibinfo{person}{John Schulman}, \bibinfo{person}{Filip Wolski}, \bibinfo{person}{Prafulla Dhariwal}, \bibinfo{person}{Alec Radford}, {and} \bibinfo{person}{Oleg Klimov}.} \bibinfo{year}{2017}\natexlab{}.
\newblock \showarticletitle{Proximal policy optimization algorithms}.
\newblock \bibinfo{journal}{\emph{arXiv preprint arXiv:1707.06347}} (\bibinfo{year}{2017}).
\newblock


\bibitem[Shen et~al\mbox{.}(2020)]%
        {acdne}
\bibfield{author}{\bibinfo{person}{Xiao Shen}, \bibinfo{person}{Quanyu Dai}, \bibinfo{person}{Fu-lai Chung}, \bibinfo{person}{Wei Lu}, {and} \bibinfo{person}{Kup-Sze Choi}.} \bibinfo{year}{2020}\natexlab{}.
\newblock \showarticletitle{Adversarial deep network embedding for cross-network node classification}. In \bibinfo{booktitle}{\emph{Proceedings of the AAAI conference on artificial intelligence}}, Vol.~\bibinfo{volume}{34}. \bibinfo{pages}{2991--2999}.
\newblock


\bibitem[Shen et~al\mbox{.}(2023)]%
        {shen2023domain}
\bibfield{author}{\bibinfo{person}{Xiao Shen}, \bibinfo{person}{Shirui Pan}, \bibinfo{person}{Kup-Sze Choi}, {and} \bibinfo{person}{Xi Zhou}.} \bibinfo{year}{2023}\natexlab{}.
\newblock \showarticletitle{Domain-adaptive message passing graph neural network}.
\newblock \bibinfo{journal}{\emph{Neural Networks}}  \bibinfo{volume}{164} (\bibinfo{year}{2023}), \bibinfo{pages}{439--454}.
\newblock


\bibitem[Sun and Fan(2024)]%
        {sun2024mmd}
\bibfield{author}{\bibinfo{person}{Yan Sun} {and} \bibinfo{person}{Jicong Fan}.} \bibinfo{year}{2024}\natexlab{}.
\newblock \showarticletitle{Mmd graph kernel: Effective metric learning for graphs via maximum mean discrepancy}. In \bibinfo{booktitle}{\emph{The Twelfth International Conference on Learning Representations}}.
\newblock


\bibitem[Swetha et~al\mbox{.}(2024)]%
        {swetha2024x}
\bibfield{author}{\bibinfo{person}{Sirnam Swetha}, \bibinfo{person}{Jinyu Yang}, \bibinfo{person}{Tal Neiman}, \bibinfo{person}{Mamshad~Nayeem Rizve}, \bibinfo{person}{Son Tran}, \bibinfo{person}{Benjamin Yao}, \bibinfo{person}{Trishul Chilimbi}, {and} \bibinfo{person}{Mubarak Shah}.} \bibinfo{year}{2024}\natexlab{}.
\newblock \showarticletitle{X-former: Unifying contrastive and reconstruction learning for mllms}.
\newblock \bibinfo{journal}{\emph{arXiv preprint arXiv:2407.13851}} (\bibinfo{year}{2024}).
\newblock


\bibitem[Tang et~al\mbox{.}(2008)]%
        {tang2008arnetminer}
\bibfield{author}{\bibinfo{person}{Jie Tang}, \bibinfo{person}{Jing Zhang}, \bibinfo{person}{Limin Yao}, \bibinfo{person}{Juanzi Li}, \bibinfo{person}{Li Zhang}, {and} \bibinfo{person}{Zhong Su}.} \bibinfo{year}{2008}\natexlab{}.
\newblock \showarticletitle{Arnetminer: extraction and mining of academic social networks}. In \bibinfo{booktitle}{\emph{Proceedings of the 14th ACM SIGKDD international conference on Knowledge discovery and data mining}}. \bibinfo{pages}{990--998}.
\newblock


\bibitem[Tomar et~al\mbox{.}(2023)]%
        {tomar2023tesla}
\bibfield{author}{\bibinfo{person}{Devavrat Tomar}, \bibinfo{person}{Guillaume Vray}, \bibinfo{person}{Behzad Bozorgtabar}, {and} \bibinfo{person}{Jean-Philippe Thiran}.} \bibinfo{year}{2023}\natexlab{}.
\newblock \showarticletitle{Tesla: Test-time self-learning with automatic adversarial augmentation}. In \bibinfo{booktitle}{\emph{Proceedings of the IEEE/CVF conference on computer vision and pattern recognition}}. \bibinfo{pages}{20341--20350}.
\newblock


\bibitem[Van~der Maaten and Hinton(2008)]%
        {t-sne}
\bibfield{author}{\bibinfo{person}{Laurens Van~der Maaten} {and} \bibinfo{person}{Geoffrey Hinton}.} \bibinfo{year}{2008}\natexlab{}.
\newblock \showarticletitle{Visualizing data using t-SNE.}
\newblock \bibinfo{journal}{\emph{Journal of machine learning research}} \bibinfo{volume}{9}, \bibinfo{number}{11} (\bibinfo{year}{2008}).
\newblock


\bibitem[Vaswani et~al\mbox{.}(2017)]%
        {vaswani2017attention}
\bibfield{author}{\bibinfo{person}{Ashish Vaswani}, \bibinfo{person}{Noam Shazeer}, \bibinfo{person}{Niki Parmar}, \bibinfo{person}{Jakob Uszkoreit}, \bibinfo{person}{Llion Jones}, \bibinfo{person}{Aidan~N Gomez}, \bibinfo{person}{{\L}ukasz Kaiser}, {and} \bibinfo{person}{Illia Polosukhin}.} \bibinfo{year}{2017}\natexlab{}.
\newblock \showarticletitle{Attention is all you need}.
\newblock \bibinfo{journal}{\emph{Advances in neural information processing systems}}  \bibinfo{volume}{30} (\bibinfo{year}{2017}).
\newblock


\bibitem[Veli{\v{c}}kovi{\'c} et~al\mbox{.}(2017)]%
        {gat}
\bibfield{author}{\bibinfo{person}{Petar Veli{\v{c}}kovi{\'c}}, \bibinfo{person}{Guillem Cucurull}, \bibinfo{person}{Arantxa Casanova}, \bibinfo{person}{Adriana Romero}, \bibinfo{person}{Pietro Lio}, {and} \bibinfo{person}{Yoshua Bengio}.} \bibinfo{year}{2017}\natexlab{}.
\newblock \showarticletitle{Graph attention networks}.
\newblock \bibinfo{journal}{\emph{arXiv preprint arXiv:1710.10903}} (\bibinfo{year}{2017}).
\newblock


\bibitem[Wu et~al\mbox{.}(2020)]%
        {udagcn}
\bibfield{author}{\bibinfo{person}{Man Wu}, \bibinfo{person}{Shirui Pan}, \bibinfo{person}{Chuan Zhou}, \bibinfo{person}{Xiaojun Chang}, {and} \bibinfo{person}{Xingquan Zhu}.} \bibinfo{year}{2020}\natexlab{}.
\newblock \showarticletitle{Unsupervised domain adaptive graph convolutional networks}. In \bibinfo{booktitle}{\emph{Proceedings of the web conference 2020}}. \bibinfo{pages}{1457--1467}.
\newblock


\bibitem[Wu et~al\mbox{.}(2024)]%
        {wu2024graph}
\bibfield{author}{\bibinfo{person}{Man Wu}, \bibinfo{person}{Xin Zheng}, \bibinfo{person}{Qin Zhang}, \bibinfo{person}{Xiao Shen}, \bibinfo{person}{Xiong Luo}, \bibinfo{person}{Xingquan Zhu}, {and} \bibinfo{person}{Shirui Pan}.} \bibinfo{year}{2024}\natexlab{}.
\newblock \showarticletitle{Graph Learning under Distribution Shifts: A Comprehensive Survey on Domain Adaptation, Out-of-distribution, and Continual Learning}.
\newblock \bibinfo{journal}{\emph{arXiv preprint arXiv:2402.16374}} (\bibinfo{year}{2024}).
\newblock


\bibitem[Wu et~al\mbox{.}(2022)]%
        {eerm}
\bibfield{author}{\bibinfo{person}{Qitian Wu}, \bibinfo{person}{Hengrui Zhang}, \bibinfo{person}{Junchi Yan}, {and} \bibinfo{person}{David Wipf}.} \bibinfo{year}{2022}\natexlab{}.
\newblock \showarticletitle{Handling distribution shifts on graphs: An invariance perspective}.
\newblock \bibinfo{journal}{\emph{arXiv preprint arXiv:2202.02466}} (\bibinfo{year}{2022}).
\newblock


\bibitem[Xiao et~al\mbox{.}(2024)]%
        {semigcl}
\bibfield{author}{\bibinfo{person}{Jiaren Xiao}, \bibinfo{person}{Quanyu Dai}, \bibinfo{person}{Xiao Shen}, \bibinfo{person}{Xiaochen Xie}, \bibinfo{person}{Jing Dai}, \bibinfo{person}{James Lam}, {and} \bibinfo{person}{Ka-Wai Kwok}.} \bibinfo{year}{2024}\natexlab{}.
\newblock \showarticletitle{Semi-supervised domain adaptation on graphs with contrastive learning and minimax entropy}.
\newblock \bibinfo{journal}{\emph{Neurocomputing}}  \bibinfo{volume}{580} (\bibinfo{year}{2024}), \bibinfo{pages}{127469}.
\newblock


\bibitem[Xu et~al\mbox{.}(2018)]%
        {gin}
\bibfield{author}{\bibinfo{person}{Keyulu Xu}, \bibinfo{person}{Weihua Hu}, \bibinfo{person}{Jure Leskovec}, {and} \bibinfo{person}{Stefanie Jegelka}.} \bibinfo{year}{2018}\natexlab{}.
\newblock \showarticletitle{How powerful are graph neural networks?}
\newblock \bibinfo{journal}{\emph{arXiv preprint arXiv:1810.00826}} (\bibinfo{year}{2018}).
\newblock


\bibitem[Yan et~al\mbox{.}(2017)]%
        {yan2017mind}
\bibfield{author}{\bibinfo{person}{Hongliang Yan}, \bibinfo{person}{Yukang Ding}, \bibinfo{person}{Peihua Li}, \bibinfo{person}{Qilong Wang}, \bibinfo{person}{Yong Xu}, {and} \bibinfo{person}{Wangmeng Zuo}.} \bibinfo{year}{2017}\natexlab{}.
\newblock \showarticletitle{Mind the class weight bias: Weighted maximum mean discrepancy for unsupervised domain adaptation}. In \bibinfo{booktitle}{\emph{Proceedings of the IEEE conference on computer vision and pattern recognition}}. \bibinfo{pages}{2272--2281}.
\newblock


\bibitem[Yang et~al\mbox{.}(2023)]%
        {yang2023diffusion}
\bibfield{author}{\bibinfo{person}{Ling Yang}, \bibinfo{person}{Zhilong Zhang}, \bibinfo{person}{Yang Song}, \bibinfo{person}{Shenda Hong}, \bibinfo{person}{Runsheng Xu}, \bibinfo{person}{Yue Zhao}, \bibinfo{person}{Wentao Zhang}, \bibinfo{person}{Bin Cui}, {and} \bibinfo{person}{Ming-Hsuan Yang}.} \bibinfo{year}{2023}\natexlab{}.
\newblock \showarticletitle{Diffusion models: A comprehensive survey of methods and applications}.
\newblock \bibinfo{journal}{\emph{ACM computing surveys}} \bibinfo{volume}{56}, \bibinfo{number}{4} (\bibinfo{year}{2023}), \bibinfo{pages}{1--39}.
\newblock


\bibitem[You et~al\mbox{.}(2023)]%
        {mfrreg}
\bibfield{author}{\bibinfo{person}{Yuning You}, \bibinfo{person}{Tianlong Chen}, \bibinfo{person}{Zhangyang Wang}, {and} \bibinfo{person}{Yang Shen}.} \bibinfo{year}{2023}\natexlab{}.
\newblock \showarticletitle{Graph domain adaptation via theory-grounded spectral regularization}. In \bibinfo{booktitle}{\emph{The eleventh international conference on learning representations}}.
\newblock


\bibitem[Zhang et~al\mbox{.}(2024)]%
        {zhang2024collaborate}
\bibfield{author}{\bibinfo{person}{Zhen Zhang}, \bibinfo{person}{Meihan Liu}, \bibinfo{person}{Anhui Wang}, \bibinfo{person}{Hongyang Chen}, \bibinfo{person}{Zhao Li}, \bibinfo{person}{Jiajun Bu}, {and} \bibinfo{person}{Bingsheng He}.} \bibinfo{year}{2024}\natexlab{}.
\newblock \showarticletitle{Collaborate to Adapt: Source-Free Graph Domain Adaptation via Bi-directional Adaptation}. In \bibinfo{booktitle}{\emph{Proceedings of the ACM on Web Conference 2024}}. \bibinfo{pages}{664--675}.
\newblock


\end{thebibliography}
